\documentclass{article}

\PassOptionsToPackage{square,numbers}{natbib}
\usepackage[preprint]{neurips_2023}

\usepackage[utf8]{inputenc} 
\usepackage[T1]{fontenc}    
\usepackage{hyperref}       
\usepackage{url}            
\usepackage{booktabs}       
\usepackage{amsfonts}       
\usepackage{amsmath}
\usepackage{amssymb}
\usepackage{mathtools}
\usepackage{nicefrac}       
\usepackage{microtype}      
\usepackage[colorinlistoftodos]{todonotes}
\usepackage{paralist}
\usepackage{xcolor}         
\usepackage{xspace}
\usepackage{tikz}
\usepackage{caption}
\usepackage{subcaption}
\usepackage{upgreek}

\newboolean{showcomments}
\setboolean{showcomments}{false}

\newcommand{\customtodo}[2][]{
\ifshowcomments
\textcolor{#1}{#2}
\fi
}

\newcommand{\MM}[2][]{
\customtodo[cyan]{\textbf{MM:} #2\xspace}
}

\newcommand{\KL}[2][]{
\customtodo[orange]{\textbf{KL:} #2\xspace}
}

\newtheorem{dfn}{Definition}
\newtheorem{proof}{Proof}
\newtheorem{proofsketch}{Proof Sketch}
\newtheorem{assumption}{Assumption}
\newtheorem{remark}{Remark}
\newtheorem{problem}{Problem}
\newtheorem{thm}{Theorem}
\newtheorem{lemma}{Lemma}

\DeclareMathOperator*{\argmax}{arg\,max}

\newcommand{\props}{\ensuremath{\mathcal{P}}\xspace}

\newcommand{\done}{\texttt{done}\xspace}

\newcommand{\exec}{\ensuremath{x}\xspace}
\newcommand{\goal}{\ensuremath{\mathcal{G}}\xspace}
\newcommand{\pmult}{\ensuremath{C_p}\xspace}
\newcommand{\Q}{\ensuremath{\bar{Q}}\xspace}

\newcommand{\bulletsafety}{\texttt{Bullet-Safety-Gym}\xspace}

\title{Safety-Aware Task Composition for Discrete and Continuous Reinforcement Learning}

\author{%
  Kevin Leahy\thanks{The first two authors contributed equally.\\DISTRIBUTION STATEMENT A. Approved for public release. Distribution is unlimited.This material is based upon work supported by the Under Secretary of Defense for Research and Engineering under Air Force Contract No. FA8702-15-D-0001. Any opinions, findings, conclusions or recommendations expressed in this material are those of the author(s) and do not necessarily reflect the views of the Under Secretary of Defense for Research and Engineering.}, Makai Mann\footnotemark[1], Zachary Serlin \\
  MIT Lincoln Laboratory\\
  Lexington, MA 02421 \\
  \texttt{\{kevin.leahy,makai.mann,zachary.serlin\}@ll.mit.edu} \\
}

\begin{document}

\maketitle

\begin{abstract}
Compositionality is a critical aspect of scalable system design. Reinforcement learning (RL) has recently shown substantial success in task learning, but has only recently begun to truly leverage composition. In this paper, we focus on Boolean composition of learned tasks as opposed to functional or sequential composition. Existing Boolean composition for RL focuses on reaching a satisfying absorbing state in environments with discrete action spaces, but does not support composable safety (i.e., avoidance) constraints. We advance the state of the art in Boolean composition of learned tasks with three contributions: 
\begin{inparaenum}[i)]
\item introduce two distinct notions of safety in this framework; 
\item show how to enforce either safety semantics, prove correctness (under some assumptions), and analyze the trade-offs between the two safety notions; and 
\item extend Boolean composition from discrete action spaces to continuous action spaces. 
\end{inparaenum}
We demonstrate these techniques using modified versions of value iteration in a grid world, Deep Q-Network (DQN) in a grid world with image observations, and Twin Delayed DDPG (TD3) in a continuous-observation and continuous-action Bullet physics environment. We believe that these contributions advance the theory of safe reinforcement learning
by allowing zero-shot composition of policies satisfying safety properties. 
\end{abstract}

\section{Introduction}

Recent advances have established reinforcement learning (RL) as a powerful tool for human-level performance in games~\cite{schrittwieser2020}, robotics~\cite{ibarz2021}, and other fields~\cite{arulkumaran2017}. However, there are still many technical hurdles to deploying these algorithms in real-world scenarios. For many of these approaches, millions of samples are required before the desired behavior is achieved. Additionally, there are concerns about side effects, reward hacking, and transparency~\cite{amodei2016}. Transfer learning~\cite{transfer_learning}, and in particular task composition~\cite{boolean-task-algebra}, has emerged as a promising method for learning simple task primitives and composing them in a zero-shot manner to perform more complex behaviors. Composing tasks in this way assures that if desired behavior is achieved at the level of task primitives, then desired behavior will be achieved when tasks are composed.

Prior work in composition has focused primarily on reachability problems, in the form of stochastic shortest paths~\cite{boolean-task-algebra,tasse2022,skill-machines}. However, another important property is safety. That is, avoiding undesirable states on the way to reaching a final state. For example, in an autonomous driving scenario, if a truck is laden with hazardous materials, it may need to avoid certain neighborhoods, whereas, if it is not carrying such materials, it does not need to avoid those routes. Thus, it is desirable to have a policy for achieving a shortest path, as well as a policy for avoiding forbidden neighborhoods. At runtime, the shortest path policy can be run on its own, or it can be composed with the safety policy as well. 
In this work, we present a method for composing simple policies to satisfy complex behaviors with no additional training that also include safety considerations. Safety-aware learning has become of great interest in recent years because it is necessary when deploying learning-based systems in safety-critical applications (e.g., autonomous driving). 
These concepts have long existed in the controls community \cite{cbfs_overview}, but have recently been applied to RL as well \cite{rl-shielding,hierarchical-potential-rewards,safe_learning_chuchu}.

This work combines components from both the safety-aware learning and task composition communities to create safety-aware task composition. There is a fundamental trade-off between safety (avoiding unsafe states) and liveness (reaching a set of goal states). This work generates policies that prioritize liveness, but are safety-aware and will minimally violate safety conditions when necessary. We do this by learning safe policies for basic reach-avoid tasks with a reward structure that incentivizes reaching the desired goal region, but also penalizes paths that enter undesirable regions.

\paragraph{Contributions:} The main contribution of this paper is a method for generating and composing multiple safety-aware policies using Q-learning approaches that can be combined at deployment using the presented safety-aware Boolean task algebra formulation. Specifically, (1) we present two new semantics to create safety-aware Boolean task algebra that encode safety constraints as (a) minimally entering states that are not explicitly goal states (by count) and (b) minimally entering states that are prioritized as bad states; (2) we show how to enforce either safety semantics, prove correctness under some assumptions, and analyze the trade-offs between the two safety notions; and (3) we extend Boolean composition from discrete action spaces to continuous action spaces. We demonstrate these techniques using modified versions of value iteration in a grid world, Deep Q-Network (DQN) in a grid world with image observations, and Twin Delayed DDPG (TD3) in a continuous-observation and continuous-action Bullet physics simulation environment.

\paragraph{Related work:} 

This paper is most closely related to \citet{boolean-task-algebra} and its extensions \cite{tasse2022,skill-machines}. In \cite{boolean-task-algebra}, the authors present a Boolean task algebra for composing simple discrete Q-tables. We extend this work in two ways: 1) by reformulating the task algebra to include two forms of safety, and 2) by extending the formulation and guarantees to continuous action spaces. 
There are also several approaches to synthesizing rewards that enable both safety and reachability according to hierarchical structures~\cite{hierarchical-potential-rewards} or specification languages~\cite{jothimurugan2019,jothimurugan2021compositional}. However, these approaches are not designed to perform compositionally. That is, to combine two tasks, a new policy needs to be trained.
Further, there has been an effort to address more general forms of policy composition~\cite{adamczyk2023compositionality}. Such work provides bounded optimality on the arbitrary composition of tasks, but does not provide guarantees on exact zero-shot composition.

Similarly, there are a number of approaches that use abstract graphs to encode both rewards and safety requirements (e.g., in~\cite{science_composition}). These approaches are generally referred to as shielding in RL~\cite{rl-shielding}, and can be very effective at producing safe policies by generally filtering a policies actions such that only safe actions are allowed by the agent and a negative reward is imposed if the filter is activated. Decomposition approaches typically generate a reward automata (or reward machine) encoding using simple dense rewards as guards on an automata that encodes a temporal logic specification (e.g., \cite{balakrishnan2022}). 
Such decomposition methods provide tools for achieving safety and reachability, while breaking an end-to-end task into smaller problems, but they do not allow composition. That is, they can take a monolithic task and break it into smaller learning problems, whereas our work focuses on building up complex capabilities at deployment from a general set of simple pre-trained tasks.
\section{Problem formulation}

\subsection{Markov decision processes and reinforcement learning}\label{sec:mdps}

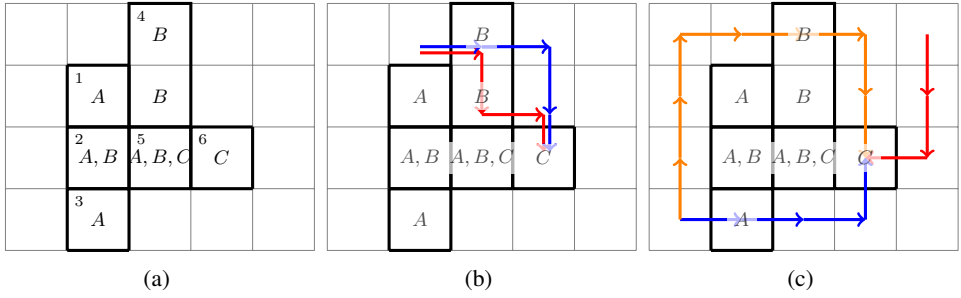
\begin{figure}
\begin{subfigure}[b]{0.30\textwidth}
    \centering
    \resizebox{\linewidth}{!}{
    \begin{tikzpicture}
    \draw[step=1.0cm,color=gray] (0,0) grid (5,4);
    \draw[step=1.0cm,color=black,line width=.5mm] (1,0) grid (2,1);
    \draw[step=1.0cm,color=black,line width=.5mm] (1,1) grid (2,2);
    \draw[step=1.0cm,color=black,line width=.5mm] (1,2) grid (2,3);
    \draw[step=1.0cm,color=black,line width=.5mm] (2,1) grid (3,2);
    \draw[color=black,line width=.5mm] (2,2) rectangle (3,4);
    \draw[step=1.0cm,color=black,line width=.5mm] (3,1) grid (4,2);

    \draw (1,3) node[anchor=north west, font=\scriptsize] {\(1\)};
    \draw (1,2) node[anchor=north west, font=\scriptsize] {\(2\)};
    \draw (1,1) node[anchor=north west, font=\scriptsize] {\(3\)};
    \draw (2,4) node[anchor=north west, font=\scriptsize] {\(4\)};
    \draw (2,2) node[anchor=north west, font=\scriptsize] {\(5\)};
    \draw (3,2) node[anchor=north west, font=\scriptsize] {\(6\)};
    
    \node[font=\small] at (1.5,2.5) {\(A\)};
    \node[font=\small] at (1.5,1.5) {\(A,B\)};
    \node[font=\small] at (1.5,0.5) {\(A\)};
    \node[font=\small] at (2.5,3.5) {\(B\)};
    \node[font=\small] at (2.5,2.5) {\(B\)};
    \node[font=\small] at (2.5,1.5) {\(A,B,C\)};
    \node[font=\small] at (3.5,1.5) {\(C\)};
    
    \end{tikzpicture}}\caption{}\label{fig:empty_env}

\end{subfigure}
  \begin{subfigure}[b]{0.30\textwidth}
    \centering
    \resizebox{\linewidth}{!}{
    \begin{tikzpicture}
    \draw[step=1.0cm,color=gray] (0,0) grid (5,4);
    \draw[step=1.0cm,color=black,line width=.5mm] (1,0) grid (2,1);
    \draw[step=1.0cm,color=black,line width=.5mm] (1,1) grid (2,2);
    \draw[step=1.0cm,color=black,line width=.5mm] (1,2) grid (2,3);
    \draw[step=1.0cm,color=black,line width=.5mm] (2,1) grid (3,2);
    \draw[color=black,line width=.5mm] (2,2) rectangle (3,4);
    \draw[step=1.0cm,color=black,line width=.5mm] (3,1) grid (4,2);

    \draw [->,red,line width=.5mm](1.5 , 3.2) -- (2.5, 3.2);
    \draw [->,red,line width=.5mm](2.5 , 3.2) -- (2.5, 2.2);
    \draw [->,red,line width=.5mm](2.5 , 2.2) -- (3.5, 2.2);
    \draw [->,red,line width=.5mm](3.5 , 2.2) -- (3.5, 1.6);
    
    \draw [->,blue,line width=.5mm](1.5 , 3.3) -- (2.5, 3.3);
    \draw [->,blue,line width=.5mm](2.5 , 3.3) -- (3.6, 3.3);
    \draw [->,blue,line width=.5mm](3.6 , 3.3) -- (3.6, 2.2);
    \draw [->,blue,line width=.5mm](3.6 , 2.2) -- (3.6, 1.6);
    \node[font=\small,fill=white,opacity=.7] at (1.5,2.5) {\(A\)};
    \node[font=\small,fill=white,opacity=.7] at (1.5,1.5) {\(A,B\)};
    \node[font=\small,fill=white,opacity=.7] at (1.5,0.5) {\(A\)};
    \node[font=\small,fill=white,opacity=.7] at (2.5,3.5) {\(B\)};
    \node[font=\small,fill=white,opacity=.7] at (2.5,2.5) {\(B\)};
    \node[font=\small,fill=white,opacity=.7] at (2.5,1.5) {\(A,B,C\)};
    \node[font=\small,fill=white,opacity=.7] at (3.5,1.5) {\(C\)};
    \end{tikzpicture}}\caption{}\label{fig:env_paths}
\end{subfigure}
  \begin{subfigure}[b]{0.30\textwidth}
    \centering
    \resizebox{\linewidth}{!}{
    \begin{tikzpicture}
    \draw[step=1.0cm,color=gray] (0,0) grid (5,4);
    \draw[step=1.0cm,color=black,line width=.5mm] (1,0) grid (2,1);
    \draw[step=1.0cm,color=black,line width=.5mm] (1,1) grid (2,2);
    \draw[step=1.0cm,color=black,line width=.5mm] (1,2) grid (2,3);
    \draw[step=1.0cm,color=black,line width=.5mm] (2,1) grid (3,2);
    \draw[color=black,line width=.5mm] (2,2) rectangle (3,4);
    \draw[step=1.0cm,color=black,line width=.5mm] (3,1) grid (4,2);

    \draw [->,red,line width=.5mm](4.5 , 3.5) -- (4.5, 2.5);
    \draw [->,red,line width=.5mm](4.5 , 2.5) -- (4.5, 1.5);
    \draw [->,red,line width=.5mm](4.5 , 1.5) -- (3.5, 1.5);
    
    \draw [->,blue,line width=.5mm](0.5 , 0.5) -- (1.5, 0.5);
    \draw [->,blue,line width=.5mm](1.5 , 0.5) -- (2.5, 0.5);
    \draw [->,blue,line width=.5mm](2.5 , 0.5) -- (3.5, 0.5);
    \draw [->,blue,line width=.5mm](3.5 , 0.5) -- (3.5, 1.5);
    
    \draw [->,orange,line width=.5mm](0.5 , 0.5) -- (0.5, 1.5);
    \draw [->,orange,line width=.5mm](0.5 , 1.5) -- (0.5, 2.5);
    \draw [->,orange,line width=.5mm](0.5 , 2.5) -- (0.5, 3.5);
    \draw [->,orange,line width=.5mm](0.5 , 3.5) -- (1.5, 3.5);
    \draw [->,orange,line width=.5mm](1.5 , 3.5) -- (2.5, 3.5);
    \draw [->,orange,line width=.5mm](2.5 , 3.5) -- (3.5, 3.5);
    \draw [->,orange,line width=.5mm](3.5 , 3.5) -- (3.5, 2.5);
    \draw [->,orange,line width=.5mm](3.5 , 2.5) -- (3.5, 1.5);
    
    \node[font=\small,fill=white,opacity=.7] at (1.5,2.5) {\(A\)};
    \node[font=\small,fill=white,opacity=.7] at (1.5,1.5) {\(A,B\)};
    \node[font=\small,fill=white,opacity=.7] at (1.5,0.5) {\(A\)};
    \node[font=\small,fill=white,opacity=.7] at (2.5,3.5) {\(B\)};
    \node[font=\small,fill=white,opacity=.7] at (2.5,2.5) {\(B\)};
    \node[font=\small,fill=white,opacity=.7] at (2.5,1.5) {\(A,B,C\)};
    \node[font=\small,fill=white,opacity=.7] at (3.5,1.5) {\(C\)};
    \end{tikzpicture}}\caption{}\label{fig:safe_paths}
\end{subfigure}

\caption{Example environment. \ref{fig:empty_env}: \(\props=\lbrace A,B,C\rbrace\). \(\goal\) consists of 6 regions, numbered 1--6, with borders indicated by dark lines. \ref{fig:env_paths}: Two sample paths through the environment. The associated sequence of labels for both paths is \(\lbrace\emptyset,B,\emptyset,\emptyset,C\rbrace\), despite the differing paths. \ref{fig:safe_paths}: Different types of paths that all satisfy \(\phi = C\). A \emph{pure} path in red, a \emph{minimum-violation} path in blue, and a \emph{prioritized safety} path in orange for \(\phi = \neg A\wedge C\). Note that the minimum-violation and prioritized safety paths start in the same region, but the prioritized safety path is much longer to avoid producing the label \(A\).}\label{fig:env}
\end{figure}

Let \props be a set of atomic propositions. We model an agent's environment (and its motion in the environment) as a labeled Markov decision process (MDP). An MDP is written as a tuple \(\langle \mathcal{S},\mathcal{A},\tau,R, L\rangle\), where \(\mathcal{S}\) is the state space, \(\mathcal{A}\) is the action space, 
\(\tau:\mathcal{S}\times\mathcal{A}\times\mathcal{S}\rightarrow [0,1]\) is the transition probability, \(R:\mathcal{S}\times\mathcal{A}\times\mathcal{S}\rightarrow\mathbb{R}\) is the reward function,  and \(L: \mathcal{S} \rightarrow 2^\props\) is a labeling function mapping each state to a set of atomic propositions.
\KL{Note for future reference: since the MDP is deterministic, \(\tau:\mathcal{S}\times\mathcal{A}\rightarrow\mathcal{S}\) as a transition \emph{function} is probably a better choice.}

An \emph{execution} of an MDP is a finite sequence of tuples \(\exec=\langle s_0, \emptyset \rangle, \langle s_1, l_1 \rangle, \dots\), where $s_i \in \mathcal{S}$ are states and $l_i \in 2^\props$ are labels.
We consider MDPs that only emit symbols when the label given by \(L\) changes. 
Thus, the first label in an execution is always the empty set. 
This behavior is illustrated in Fig.~\ref{fig:env_paths}. The red path and the blue path produce the same sequence of labels. Even though the red path passes through two states labeled \(B\), the second state produces \(\emptyset\), since it shares the same label as the previous state.

We further introduce the \emph{projection} of an execution \({\upharpoonright}_L:\mathcal{S}\times 2^\props\rightarrow 2^\props\), that projects an execution onto the set of associated labels. That is, for an execution \(\exec=\langle s_0, l_0 \rangle, \langle s_1, l_1 \rangle, \dots\), we write \({\upharpoonright}_L(\exec)=l_0,l_1,\ldots\). We denote the sequence of non-empty symbols from the projection as \({\upharpoonright}^+_L\). For example, if \({\upharpoonright}_L(\exec)=\emptyset,\emptyset,A,\emptyset,\lbrace A,B\rbrace,\emptyset,B\), then \({\upharpoonright}^+_L(\exec)=A,\lbrace A,B\rbrace,B\). Let \(\vert{\upharpoonright}_L\vert\) and \(\vert{\upharpoonright}^+_L\vert\) denote the length of each of these projections. 

The labeling function of an MDP induces a set \(\goal \in 2^\mathcal{S}\), consisting of a set of disjoint, connected subsets of \(\mathcal{S}\). We assume each region \(g \in \goal\) is mapped to a non-empty label by \(L\). We refer to these as \emph{goals}. Any state not belonging to any \(g \in \goal\) is unlabeled and maps to the empty set. For \(p\in \props\), if \(p \in L(g)\), we call \(g\) a \(p\)-region and otherwise we call it a \(\neg p\)-region. 
A given proposition \(p\) induces a partition over the regions in \goal, with \(\goal_p\coloneqq\lbrace g\in\goal\mid p\in L(g)\rbrace\) and \(\goal_{\neg p}\coloneqq\lbrace g\in\goal\mid p\not\in L(g)\rbrace\), such that \(\goal_p\cap\goal_{\neg p}=\emptyset\). That is, \(\goal_p\) contains all goals that satisfy \(p\) and \(\goal_{\neg p}\) contains all goals that do not satisfy \(p\). 
Fig.~\ref{fig:empty_env} illustrates a labeled environment and shows the differences between \props and \goal. 

\begin{remark}
    In this work, each goal \(g\in\goal\) is a \emph{set} of states. In~\citet{boolean-task-algebra}, this same term refers to individual states. While the definition we use is slightly different, functionally, these two definitions behave similarly.
\end{remark}

We can inductively define Boolean formulas over \props as
\begin{equation}
    \phi\coloneqq p \mid \neg \phi \mid \phi_1 \wedge \phi_2 \mid \phi_1 \vee \phi_2\:,
\end{equation}
where \(p\in\props\) is a proposition; \(\phi\) is a formula; and \(\neg\), \(\wedge\), and \(\vee\) are the standard Boolean operations of negation, conjunction, and disjunction, respectively.

For a given Boolean formula \(\phi\) over \(\props\), we say that an execution \(\exec\) of an MDP satisfies the formula (written \(\exec\models\phi\)) if the last element of \({\upharpoonright}^+_L(\exec)\) satisfies the formula (in the Boolean sense). E.g., if the associated task is \(A\), as long as the last element of \({\upharpoonright}^+_L(\exec)\) contains the symbol \(A\), then the task is satisfied. In other words, if an execution \exec terminates somewhere in \(\goal_p\), we say the task is satisfied.

\KL{Boolean formula over props defines a reward structure over an MDP. We call this MDP with its associated reward a \emph{task}. If there is room in the paper, include a remark that says essentially, goals are confusing. They are regions we can go to that may or may not satisfy a given task. When executing a policy, we max over the goals. So we expect that goals that satisfy the task are the ones that are selected by the policy.}


\subsection{Safety properties}\label{sec:safety}
\MM{Does it make sense for this and the following sections to still be in background?}

Previous work~\cite{james2006analysis,boolean-task-algebra,composing-value-functions} has focused on \emph{proper} policies, that is, policies that are guaranteed to reach an absorbing set. In this work, we want proper policies that have additional properties. Namely, they should avoid passing through unsafe (or undesired) states. We consider three sub-classes of proper paths in an MDP that we would like to achieve: 
\begin{enumerate}
    \item \emph{Pure} paths -- proper paths that do not produce any other symbols;
    \item \emph{Minimum-violation} paths -- proper paths that produce the minimum number of other symbols; and
    \item \emph{Prioritized safety} paths -- proper paths that completely avoid certain regions, and produce a minimum-violation path over the remaining regions.
\end{enumerate}

\paragraph{Motivation}
These safe path definitions are inspired by temporal logic (TL) planning. To accomplish a task specified in Linear Temporal Logic~\cite{temporal-logic}, an agent must satisfy Boolean formulae to take transitions in a (B{\"u}chi) automata. However, satisfying a \emph{different} formula first could lead to an unintended transition. Please see \citet{calin-book} for details on planning under TL constraints.

Intuitively, a pure path is the most desirable. Such paths terminate at the desired states and produce no extraneous symbols. Minimum-violation paths prioritize reachability goals, while minimizing extraneous symbols as much as possible. Prioritized safety paths, on the other hand, place higher weight on avoiding certain symbols. Consider \(\phi \coloneqq \neg A \wedge C\). If there is no pure path to \(C\), minimum violation would take the path that produces the fewest non-satisfying symbols, whereas prioritized safety would take a longer path to avoid producing \(A\) symbols as depicted in Fig.~\ref{fig:safe_paths}. If a pure path does not exist, a prioritized safety path may sacrifice reachability in order to satisfy safety goals. Thus, minimum-violation and prioritized safety represent different user priorities on task completion vs. safety.
To generate policies that achieve such paths, we extend the task algebra from~\cite{boolean-task-algebra} to include safety properties. We define two separate semantics for safety with solutions to satisfy each type of safety property in the sequel, and discuss trade-offs for each type of safety in Sec.~\ref{sec:policies}.
\MM{Although this sounds nice and intuitive, I'm not sure we can claim that. Once chattering starts I'm not positive it will stay safe regardless. In fact, if there's no safe path the optimal path is probably to terminate as soon as possible.}

Formally, we define these paths with respect to a Boolean formula \(\phi\) over \props as follows.
\begin{dfn}[Pure path] An execution \exec produces a \emph{pure} path if \(\vert{\upharpoonright}^+_L(x)\vert=1\) and \(\exec\models\phi\).
\end{dfn}

\begin{dfn}[Minimum-violation path]\label{def:min_viol} An execution \exec produces a \emph{minimum-violation} path if \(\vert{\upharpoonright}^+_L(x)\vert>1\) and \(\exec\models\phi\), and there is no execution \(\exec'\) such that \(\vert{\upharpoonright}^+_L(x')\vert<\vert{\upharpoonright}^+_L(x)\vert\).
\end{dfn}

\KL{What if we call this safety path, and the next definition is prioritized safety.}\MM{Made this change. Could you check for consistency?}
\begin{dfn}[Safety path]\label{def:priority} An execution \(\exec\) produces a \emph{safety} path if, for some bad formula \(\phi_B\), \(\exec\models\phi\) and no finite prefix of \(x\) satisfies \(\phi_B\).
\end{dfn}

\KL{Need to figure out def 3 vs. using a formula, for example. Easiest solution to inconsistency around negated tasks is to assume all formulas are in NNF. And then note that it's not technically required in min-violation?}

\begin{dfn}[Prioritized safety path]\label{def:prop_min_viol} An execution \exec produces a \emph{prioritized safety path} if it is a prioritized safety path, and there is no execution \(\exec'\) that is also a prioritized safety path, such that \(\vert{\upharpoonright}^+_L(x')\vert<\vert{\upharpoonright}^+_L(x)\vert\).
\end{dfn}

\subsection{Problem definition}

\begin{assumption}\label{ass:policy}
For each task \(\mathcal{M}\), there exists an optimal policy \(\pi_\mathcal{M}\), and associated Q- and value-functions, \(Q_\mathcal{M}\) and \(V_\mathcal{M}\), respectively. We are agnostic to how these policies are produced. I.e., such a policy can be found via dynamic programming techniques such as policy or value iteration, as well as by reinforcement learning techniques.
\end{assumption} 

Assumption~\ref{ass:policy} essentially states that our focus here is not on learning, per se, but rather on how to apply the results that learning a task provides to us. 
We are now ready to formally state the problem under consideration. We have one problem definition for each version of our safety semantics.

\begin{problem}\label{prob:min-viol}
Given a set of tasks \(\{\mathcal{M}\}\), define a Boolean Algebra for composing those tasks, such that safety semantics in Def.~\ref{def:min_viol} are enforced.
\end{problem}

\begin{problem}\label{prob:priority}
Given a set of tasks \(\{\mathcal{M}\}\), define a Boolean Algebra for composing those tasks, such that safety semantics in Def.~\ref{def:prop_min_viol} are enforced.
\end{problem}

\subsection{Boolean task algebra}\label{sec:task-algebra}
Here, we briefly summarize the contributions of \citet{boolean-task-algebra}, which we will build upon. Please see their paper for more details. The guarantees from this paper apply to reachability tasks and deterministic MDPs (although the latter can be relaxed in practice).

Let a set of tasks \(\{\mathcal{M}\}\) be a collection of MDPs which differ only in the reward function, \(R\), on their \emph{terminal} states \(\mathcal{G}\subseteq\mathcal{S}\).
To combine tasks in a Boolean fashion, the authors of~\cite{boolean-task-algebra} define the notion of extended reward functions and extended Q-value functions, as written below.

\begin{dfn}[Def. 2 of \cite{boolean-task-algebra}]
 The extended reward function \(\bar{r}:\mathcal{S}\times\mathcal{G}\times\mathcal{A}\rightarrow\mathbb{R}\) is defined as
 \begin{equation}
\bar{r}(s,g,a) =
\begin{cases}
\bar{r}_{MIN} &\mbox{if } g\neq s \mbox{ and } s\in\mathcal{G}\\
r(s,a) & \mbox{otherwise}\:,
\end{cases}
\end{equation}
where \(\bar{r}_{MIN}\) is a large penalty term.
\end{dfn}

\begin{dfn}[Def. 3 of \cite{boolean-task-algebra}]
An extended Q-value function \(\bar{Q}:\mathcal{S}\times\mathcal{G}\times\mathcal{A}\rightarrow\mathbb{R}\) is defined as
\begin{equation}
\bar{Q}(s,g,a)=\bar{r}(s,g,a)+\sum_{s'\in \mathcal{S}}\tau(s,a,s')\bar{V}^{\bar{\pi}}(s',g)\:,
\end{equation}
where \(\bar{V}^{\bar{\pi}}(s',g)\) is the value function corresponding to policy \(\bar{\pi}\) under reward \(\bar{r}\).
\end{dfn}

\begin{remark}
We note that the authors of~\cite{boolean-task-algebra} assume the reward function differs only on \emph{terminal} states, as written above. We will relax that assumption, as detailed in Sec.~\ref{sec:penalties}.
\end{remark}

\KL{It seems important to explain this distinction, as well as establishing to notion of tasks vs. goals. Maybe we should define tasks and goals separately even earlier in the paper?}
For a given task, there is a corresponding MDP with its own extended reward, as well as an associated extended Q-value function. The policy \(\bar{\pi}\) corresponding to an extended Q-value function is obtained by taking the \(\max\) over \(g\in\goal\). By explicitly including goals as inputs, extended Q-value functions maintain knowledge of optimal values when the goal corresponds to the task encoded in the extended reward, as well as when the goal does not correspond to the desired task. Intuitively, this captures how good or bad a given goal is in relation to the task. This subtlety is important for composition to function correctly.

With extended Q-value functions defined in this way, the Boolean operations of conjunction (\(\wedge\)), disjunction (\(\vee\)), and negation (\(\neg\)) can be performed over tasks in a zero-shot fashion as follows:
\begin{align}
\neg \bar{Q}^*(s,g,a) &= (\bar{Q}^*_{\mathcal{U}}(s,g,a)+\bar{Q}^*_\varnothing(s,g,a))-\bar{Q}^*(s,g,a) \label{eq:neg_op}\\
\bar{Q}^*_1\wedge\bar{Q}^*_2(s,g,a) &= \min\lbrace\bar{Q}^*_1(s,g,a),\bar{Q}^*_2(s,g,a)\rbrace \label{eq:conj_op}\\
\bar{Q}^*_1\vee\bar{Q}^*_2(s,g,a) &= \max\lbrace\bar{Q}^*_1(s,g,a),\bar{Q}^*_2(s,g,a)\rbrace\:,\label{eq:disj_op}
\end{align}
where \(\bar{Q}^*_{\mathcal{U}}\) and \(\bar{Q}^*_\varnothing\) correspond to the extended Q-value functions for the max and min over all rewards, respectively. Complete details are available in~\citet{boolean-task-algebra}.

While the Boolean task algebra defined this way is very powerful, it is designed to handle shortest path reachability problems. It cannot handle problems of avoidance. 
Thus, we introduce an extension of this framework that provides safety properties as defined in Sec.~\ref{sec:safety}.

\MM{Do you think it's obvious for people that negation relates to reachability and not just safety? E.g., what actually satisfied \(\neg A \wedge C\) at the end? That info is covered above, but not sure if we need a sentence on that for this audience.}


\section{Penalty-enforced safety}\label{sec:penalties}

To solve Probs.~\ref{prob:min-viol} \&~\ref{prob:priority}, we incorporate penalties on the production of labels that do not satisfy the current task. These include penalties for passing through regions that should be avoided, as well as for terminating in regions that are undesired. Intuitively we will craft the reward functions such that there is a hierarchy of bad behaviors to be avoided. Less bad behaviors will be taken to avoid worse behaviors whenever possible. Worse behaviors will have a larger penalty, and each increase in the penalty must be sufficiently large to preserve the ordering.
To enforce our penalty hierarchy, we use a \emph{penalty multiplier} \pmult. This multiplier is sufficiently large to penalize behaviors such as early termination or passing through unsafe regions. We present a method for deriving a sufficiently large value of \pmult in the Supplementary Material.


\begin{assumption} Note that our theoretical results rely on the MDP being deterministic, thus every state action pair has a deterministic transition to the next state. This can easily be relaxed in practice.
\end{assumption}



\begin{remark}
In a reinforcement learning setting we do not know \pmult exactly, because we do not know \(\tau\). However, in practice we can either estimate \(\pmult\) from interactions with the environment, set it to a very large number, or set it heuristically. Fortunately, in a deterministic MDP, \pmult has a very intuitive meaning: it is the number of steps an agent is allowed detour in order to avoid undesirable regions.
\end{remark}

\begin{table}
  \caption{Penalty hierarchy}
  \label{penalty-table}
  \centering
  \begin{tabular}{lll}
    \toprule
    Symbol     & Penalty type     & Value \\
    \midrule
    \(R_{step}\) & step & \(R_{step}\) (ex. -0.1) \\
    \(R_{badstep}\) & bad pass through & \(\pmult R_{step}\) \\
    \(R_{worst step}\) & worst pass through & \(\pmult R_{badstep}\) \\
    \(R_{bad term}\) & bad termination & \( R_{worststep}\) \\
    \(R_{worst term}\) & worst termination & \(\pmult R_{bad term}\) \\
    \bottomrule
  \end{tabular}
\end{table}

We can create a hierarchy of increasingly bad penalties by multiplying by \pmult (or an upper bound of \pmult). The hierarchy can extend indefinitely using real numbers, or until there is risk of underflow for machine representations.
Table \ref{penalty-table} depicts this penalty hierarchy. Notice that the worst pass-through penalty and the bad termination penalty are identical.
Given these penalties, we define the reward function for a given task as:
\begin{equation}\label{eq:min_viol_reward}
R_{\mathcal{M}_p}(s_i, a, g, s_{i+1}) \coloneqq  
\begin{cases}
      R_{worst term}, & \text{if}\ s_{i+1} \neq g \wedge \done \\
      R_{bad step}, & \text{if}\ s_{i+1} \neq g \wedge \neg \done \\
      R_{bad term}, & \text{if}\ s_{i+1} = g  \wedge p \not\in l_i \wedge \done \\
      R_{goal}, & \text{if}\ s_{i+1} = g  \wedge p \in l_i \wedge \done \\
      R_{step}, & \text{otherwise}\:,
\end{cases}
\end{equation}
where \done denotes the end of an episode, \(g\) and \(p\) are defined in Sec.~\ref{sec:mdps}, and \(R_{goal}\) is positive.

\subsection{Penalties for prioritized safety}\label{sec:prior_safety}
\KL{Highlight that potential chattering and having to train the policy is the price you pay for prioritized minimum violation (motivation is how much you care about different bad states)}
\KL{This is missing some context I think.}
The reward structure~\eqref{eq:min_viol_reward} makes use of the negation operator~\eqref{eq:neg_op} to enforce minimum violation semantics. For prioritized safety semantics, we instead train negated policies directly with additional penalties for \(\goal_{\neg p}\). Since we do not use a negation operator in this context, we cannot negate arbitrary Boolean formulas over tasks; however, any Boolean formula can be reduced to negation normal form (NNF). In NNF, negation only appears before literals (in this case tasks), thus we can represent any Boolean formula as conjunctions and disjunctions over positive and negated tasks.
The reward function for a negated task is defined as: 
\begin{equation}\label{eq:negation}
R_{\mathcal{M}_{\neg p}}(s_i, a, g, s_{i+1}) \coloneqq  
\begin{cases}
      R_{worst term}, & \text{if}\ s_{i+1} \neq g \wedge \done \\
      R_{worst step}, & \text{if}\ p \in l_i \wedge \neg \done \\
      R_{bad step}, & \text{if}\ s_{i+1} \neq g \wedge p \not\in l_i \wedge \neg \done \\
      R_{bad term}, & \text{if}\ s_{i+1} = g  \wedge p \in l_i \wedge \done \\
      R_{goal}, & \text{if}\ s_{i+1} = g  \wedge p \not\in l_i \wedge \done \\
      R_{step}, & \text{otherwise}
\end{cases}
\end{equation}

There are two main differences from the reward structure for a positive task. First, the conditions for obtaining \(R_{goal}\) and \(R_{bad term}\) are switched. This is because the goal is to terminate somewhere that does not satisfy \(p\). Additionally, we add a penalty for passing through any state labeled with \(p\). This penalty encourages paths that pass through regions labeled with other symbols (if necessary) rather than passing through regions containing \(p\).

This reward structure produces the correct semantics for a single negated task. In many cases, this works for composition as well. However, due to some edge cases, we cannot guarantee that (non-pure) prioritized safety paths will be taken under arbitrary compositions of negated tasks. Intuitively, this is because policies encode optimal paths for each goal, but the optimal prioritized safety path for combinations of negated tasks may not correspond to an optimal path for any of the original policies. This can result in chattering (infinite loops) in some cases. To provide formal guarantees we must make an additional assumption.
\begin{assumption}\label{ass:prioritized_safety}
We assume either:
\begin{inparaenum}[i)]
\item only a single negated policy is used in composition at a time during deployment [note this policy can be more complex than a single task (e.g., learn \(\mathcal{M}_{\neg p_1 \wedge \neg p_2}\)), but it must be learned ahead of time]; or,
\item any prioritized safety path in the environment of interest will only have to pass through a known, finite number, \(k\), of non-satisfying regions in \(\goal\) and we train an extended value function that maintains more corresponding paths.
\end{inparaenum}
\end{assumption}

See the Supplementary Details for discussion of this assumption, and details on how to train extended value functions that maintain more paths.

\subsection{Theoretical analysis and comparison of policies}\label{sec:policies}
\MM{Update this for the weakened prioritized safety semantics...}
With the reward structures described above, we wish to prove the following for both minimum-violation and prioritized safety semantics:
\begin{enumerate}
    \item If a pure path exists, the optimal policy will select it;
    \item If a pure path does not exist, the optimal policy will follow a minimum-violation (resp. prioritized safety) path; and
    \item The operations of conjunction, disjunction, and negation as defined in~\eqref{eq:neg_op}--\eqref{eq:disj_op} respect the safety semantics of the original policy.
\end{enumerate}

We now introduce theorems capturing these properties formally. Due to space constraints, proofs of these theorems can be found in the Supplementary Material.

\begin{thm}\label{thm:min_viol}
    The reward structure in~\eqref{eq:min_viol_reward} produces minimum-violation paths.
\end{thm}


\begin{thm}\label{thm:min_viol_comp}
    The composition rules in \eqref{eq:neg_op}--\eqref{eq:disj_op}, combined with tasks trained with the reward structure in~\eqref{eq:min_viol_reward}, produce behavior that is equivalent to training the negation, conjunction, and disjunction of those tasks, respectively, while maintaining minimum-violation safety semantics.
\end{thm}



\MM{Update this for the latest prioritized safety limitations...}

\begin{thm}\label{thm:priority}
    The reward structure in~\eqref{eq:min_viol_reward} for positive tasks and~\eqref{eq:negation} for negated tasks produces prioritized safety paths.
\end{thm}


\begin{thm}\label{thm:priority_comp}
    The composition rules in \eqref{eq:conj_op}--\eqref{eq:disj_op}, combined with tasks trained with the reward structure in~\eqref{eq:min_viol_reward} for positive tasks and~\eqref{eq:negation} for negated tasks, produce behavior that is equivalent to training the conjunction and disjunction of those tasks, respectively, while maintaining prioritized safety semantics under Assumption \ref{ass:prioritized_safety}.
\end{thm}



There are trade-offs associated with the choice of safety semantics. Namely, minimum-violation safety is a weaker notion of safety, treating all non-satisfying states as equally undesirable. In exchange for that weaker notion of safety, the negation operation defined in~\eqref{eq:neg_op} maintains the minimum-violation semantics. On the other hand, prioritized safety semantics have a stronger notion of safety, enforcing strict avoidance of a certain subset of \goal. To accomplish these semantics, we require Assumption \ref{ass:prioritized_safety} and train negated tasks separately. 
\MM{I commented-out a couple sentences here. Feel free to add back if you think we need them}

Additionally, the strictness of prioritized safety semantics has an impact on policies joined by conjunction and disjunction when one of the policies is a negated task. Namely, the safety property is strict enough that if there is no prioritized safety path (Def~\ref{def:prop_min_viol}) common to both policies, then there can be undesired behavior, such as chattering (infinite loops). This is because the prioritized safety semantics emphasize safety over reachability, hence the extra assumptions required in Sec.~\ref{sec:prior_safety}.
\section{Extension to continuous action spaces}\label{sec:continuous}
 
In many cases, such as robotics, it is important to support continuous action spaces. The composition method presented so far only applies for discrete action spaces. Nonetheless, we can extend composition to continuous action spaces in certain cases as well, as long as the training algorithm provides policy and value functions. Here, we show how to apply our composition to continuous action spaces.

Given two optimal extended Q-value functions, \(\Q^*_1\) and \(\Q^*_2\), with their associated optimal policies, \(\bar{\pi}^*_1\) and \(\bar{\pi}^*_2\), we can compute the policy corresponding to their conjunction as
\begin{equation}\label{eq:cont_and}
    \bar{\pi}^*_{\Q^*_1 \wedge \Q^*_2}(s,g) = 
    \begin{cases}
      \bar{\pi}^*_1(s, g) & \text{if $\Q^*_1(s,g,\bar{\pi}_1^*(s, g))\leq \Q^*_2(s,g,\bar{\pi}_2^*(s, g))$}\\
      \bar{\pi}^*_2(s, g) &  \text{otherwise}\:.
    \end{cases}  
\end{equation}

\MM{Switched to using \(\bar{\pi}^*_2\) as the argument to \(Q^*_2\) above and made a weaker claim below.}

Similarly, for disjunction, we can compute the resulting policy as
\begin{equation}\label{eq:cont_or}
    \bar{\pi}^*_{\Q^*_1 \vee \Q^*_2}(s,g) = 
    \begin{cases}
      \bar{\pi}^*_1(s, g) & \text{if $\Q^*_1(s,g,\bar{\pi}_1^*(s, g))\geq \Q^*_2(s,g,\bar{\pi}_2^*(s, g))$}\\
      \bar{\pi}^*_2(s, g) &  \text{otherwise}\:.
    \end{cases}  
\end{equation}

Negation cannot be computed directly, so we must learn a policy for each negated task. We train negated policies with minimum-violation semantics and apply them using NNF.

\begin{thm}\label{thm:continuous}
Eqs.~\eqref{eq:cont_and} \&~\eqref{eq:cont_or} 
facilitate
minimum-violation task composition.
\end{thm}

\begin{proofsketch}
Under minimum violation safety semantics, the optimal policy when composing \(\Q^*_1\) and \(\Q^*_2\) must be either \(\bar{\pi}^*_1\) or \(\bar{\pi}^*_2\). This is because minimum violation minimizes the number of non-goals appearing in an execution \(x\) and thus optimal policies must agree on \(\vert{\upharpoonright}^+_L(x)\vert\). Full details are in the Supplementary Material.
\end{proofsketch}

\MM{Actually, we think we can do this. }
This formulation does not allow for prioritized safety semantics for reasons related to Assumption \ref{ass:prioritized_safety}: the required action may not be the optimal value for either of the composed value functions. Future work may address this limitation. See the Supplementary Material for further discussion and a more detailed derivation of the policies above.


\section{Experimental evaluation}\label{sec:experiments}

We demonstrate safety-aware task composition in three environments:
\begin{enumerate}
\item a 2D static grid world with row and column observation spaces and 5 actions (each direction and stay); optimal policies obtained with value iteration\label{env:grid}
\item a 2D item collection grid world with image observations and 4 actions (each direction); optimal policies approximated by DQN\label{env:repoman}
\item a 3D physics simulation, \bulletsafety~\cite{bullet-safety-gym}, with 96D LIDAR-like observations and a (continuous) 2D force vector action space; optimal policies approximated by TD3\label{env:bullet-safety}
\end{enumerate}

Environments~\ref{env:grid} and~\ref{env:repoman} and the DQN learning infrastructure are modifications of the code from \citet{boolean-task-algebra} and \citet{composing-value-functions}. We added a collection task to \bulletsafety for Environment~\ref{env:bullet-safety} that mirrored the one in Environment~\ref{env:repoman}. The learning infrastructure used for Environment~\ref{env:bullet-safety} was a modified version of the reference TD3 implementation~\cite{td3}. All function approximation experiments were conducted with an NVIDIA Volta GPU, and tuned over three learning rates using curriculum learning, where penalties were added after the policy could successfully reach goals. We selected the best policies for the demonstrations over four random seeds. See the Supplementary Material for details and videos. 
\MM{Took out the code comment for space.} 

\begin{figure}
     \centering
     \begin{subfigure}[b]{0.3\textwidth}
         \centering   
         \includegraphics[width=\textwidth]{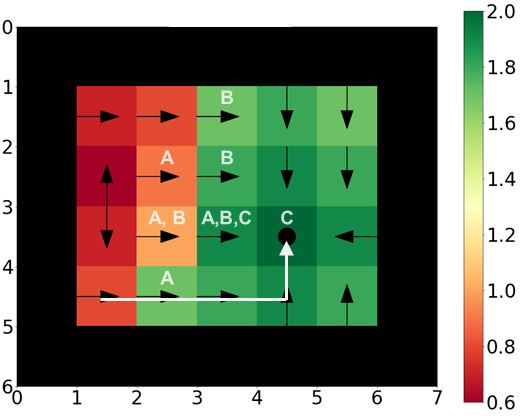}
         \caption{\(\neg A \wedge C\)}
         \label{fig:minimum_violation}
     \end{subfigure}
     \hspace{4em}
     \begin{subfigure}[b]{0.3\textwidth}
         \centering
         \includegraphics[width=\textwidth]{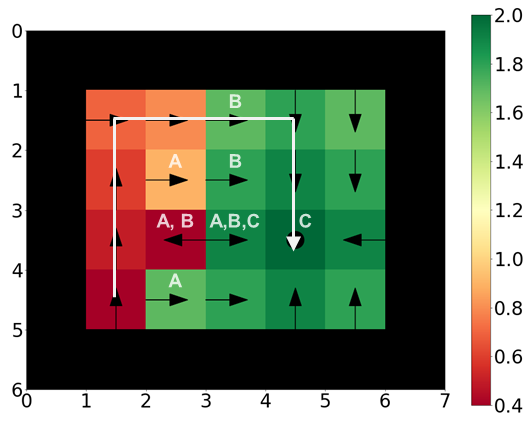}
         \caption{\(\mathit{not}\text{-}A \wedge C\)}
         \label{fig:prioritized_safety}
     \end{subfigure}

     \begin{subfigure}[b]{0.22\textwidth}
         \centering
         \includegraphics[width=\textwidth]{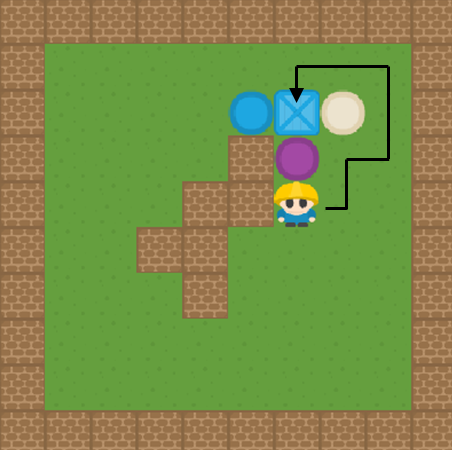}
         \caption{\(\mathit{square}\)}
         \label{fig:discrete_pure_square}
     \end{subfigure}
     \hfill
     \begin{subfigure}[b]{0.22\textwidth}
         \centering
         \includegraphics[width=\textwidth]{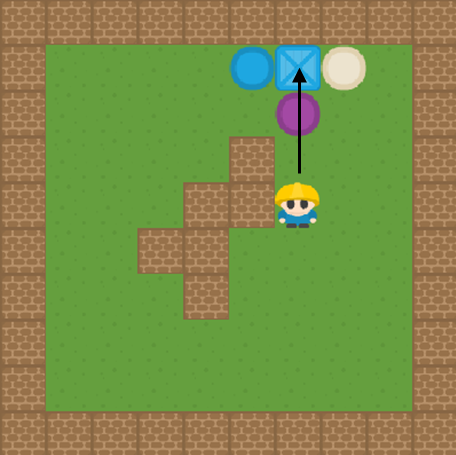}
         \caption{\(\mathit{blue} \wedge \neg \mathit{circle}\)}
         \label{fig:discrete_minimum_violation}
     \end{subfigure}
     \hfill
     \begin{subfigure}[b]{0.22\textwidth}
         \centering   
         \includegraphics[width=\textwidth]{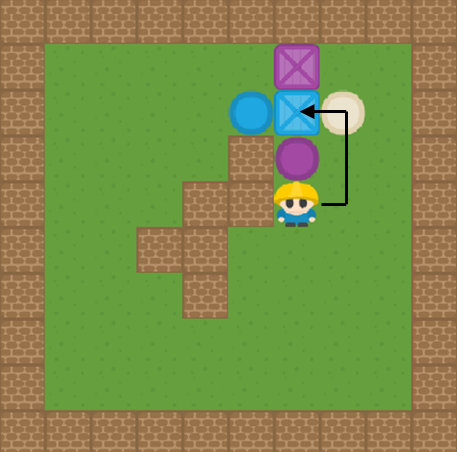}
         \caption{\(\mathit{square} \wedge \mathit{not}\text{-}\mathit{purple}\)}
         \label{fig:discrete_prioritized_safety}
     \end{subfigure}

     \begin{subfigure}[b]{0.22\textwidth}
         \centering
         \includegraphics[width=\textwidth]{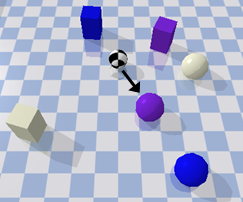}
         \caption{\(\mathit{sphere}\)}
         \label{fig:continuous_sphere}
     \end{subfigure}
     \hfill
     \begin{subfigure}[b]{0.22\textwidth}
         \centering
         \includegraphics[width=\textwidth]{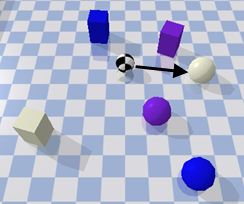}
         \caption{\(\mathit{sphere} \wedge \mathit{not}\text{-}\mathit{purple}\)}
         \label{fig:continuous_sphere_and_not-purple}
     \end{subfigure}
     \hfill
     \begin{subfigure}[b]{0.22\textwidth}
         \centering   
         \includegraphics[width=\textwidth]{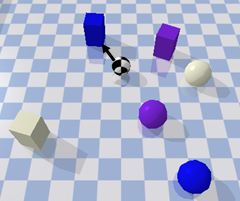}
         \caption{\(\mathit{sphere} \vee \mathit{not}\text{-}\mathit{purple}\)}
         \label{fig:continuous_sphere_or_not-purple}
     \end{subfigure}
     \hfill 
     \begin{subfigure}[b]{0.22\textwidth}
         \centering   
         \includegraphics[width=\textwidth]{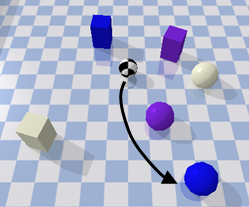}
         \caption{\(\mathit{blue} \wedge \mathit{sphere}\)}
         \label{fig:continuous_blue_and_sphere}
     \end{subfigure}
    \caption{Example optimal policies and trajectories. \textbf{Environment~\ref{env:grid}:} Color scale shows the value at the state, arrows show policy direction, and a circle denotes a stay action. \ref{fig:minimum_violation} shows a minimum violation path using analytical negation and \ref{fig:prioritized_safety} shows a prioritized safety path using a learned ``not-\(A\)" policy. Both have multiple pure paths. Note that these align with the example paths used in Fig.~\ref{fig:safe_paths}. \textbf{Environment~\ref{env:repoman}:} Fig.~\ref{fig:discrete_pure_square} shows a pure path, \ref{fig:discrete_minimum_violation} shows minimum violation using a negated task, and \ref{fig:discrete_prioritized_safety} shows a prioritized safety path using a learned negated task, ``not-\(\mathit{purple}\)". \textbf{Environment~\ref{env:bullet-safety}:} Fig.~\ref{fig:continuous_sphere} depicts a path to the nearest sphere, \ref{fig:continuous_sphere_and_not-purple} goes to the nearest sphere that is not purple, \ref{fig:continuous_sphere_or_not-purple} goes to the nearest object that is either a sphere or not purple (the blue box), and \ref{fig:continuous_blue_and_sphere} heads to the blue sphere while avoiding other goals.}
    \label{fig:continuous-env}
\end{figure}

Figs.~\ref{fig:minimum_violation}--\ref{fig:prioritized_safety} depict composed optimal policies learned with value iteration for Environment~\ref{env:grid} that highlight the different safety semantics. 
Similarly, Figs.~\ref{fig:discrete_pure_square}--\ref{fig:discrete_prioritized_safety} demonstrate different safety semantics in Environment~\ref{env:repoman}. Penalty-free Boolean task composition~\cite{boolean-task-algebra} heads straight toward a satisfying item without regard for other items in the way (not pictured). Finally, Figs.~\ref{fig:continuous_sphere}--\ref{fig:continuous_blue_and_sphere} depict our approach in Environment~\ref{env:bullet-safety}. To our knowledge, this is the first application of Boolean task composition in a continuous action space. See the Supplementary Material for additional analysis and comparisons.





\section{Limitations}
We inherit many of the same limitations of \citet{boolean-task-algebra}, including a sparse reward structure and reliance on deterministic MDPs for theoretical guarantees. We expect that reward shaping~\cite{balakrishnan2022,hierarchical-potential-rewards,reward-shaping} can be adapted to this scenario to address the former, and the latter can easily be relaxed in practice. Furthermore, Boolean task composition approaches for RL depend on the MDP being identical across tasks except the reward, and, for certain environments, not all Boolean compositions may be valid. In practice, we observed that penalties for the safety semantics made it more difficult for policies to converge. This is expected because reachability under safety constraints is complex and may require greater global reasoning to find pure or minimally violating paths. In addition, the rewards start quite negative in early exploration. We addressed this via curriculum learning, but we expect that future work in reward shaping can better address this limitation. We also note that the modularity of Boolean task composition helps identify issues during training, because each individual task can be inspected independently and trained for longer as needed.


\section{Conclusion and future work}



We have extended the theory of Boolean task composition in RL to facilitate two notions of safety constraints and support continuous action spaces. We proved correctness of the approach for optimal policies in deterministic MDPs, and demonstrated that it generalizes well to scenarios requiring function approximation. Despite some limitations, we believe that the general approach of Boolean task composition introduced by \citet{boolean-task-algebra} is significant and promising. We have addressed two such limitations by introducing safety semantics and continuous action spaces support, and recent work has demonstrated the ability to extend composition to stochastic and discounted settings~\cite{tasse2022}. We expect that future work in the community can take composition even further. This includes (potential-based) reward shaping, reducing redundancy in learning the extended value functions, and new techniques for solving the planning problem to determine which Boolean compositions to execute for more complex sequences of tasks (e.g., as defined by a temporal logic).

\paragraph{Broader impact}
Our contributions are primarily theoretical, but facilitate safer applications of RL. This advances the vision of learning agents that can execute tasks with reliable semantics and may be more explainable to humans. We acknowledge the need for further work at the intersection of computer science and psychology to support intuitive and interpretable interaction between humans and artificial agents.





\bibliography{main}

\begin{thebibliography}{24}
\providecommand{\natexlab}[1]{#1}
\providecommand{\url}[1]{\texttt{#1}}
\expandafter\ifx\csname urlstyle\endcsname\relax
  \providecommand{\doi}[1]{doi: #1}\else
  \providecommand{\doi}{doi: \begingroup \urlstyle{rm}\Url}\fi

\bibitem[Adamczyk et~al.(2023)Adamczyk, Tiomkin, and
  Kulkarni]{adamczyk2023compositionality}
J.~Adamczyk, S.~Tiomkin, and R.~Kulkarni.
\newblock Compositionality and bounds for optimal value functions in
  reinforcement learning.
\newblock \emph{arXiv preprint arXiv:2302.09676}, 2023.

\bibitem[Alshiekh et~al.(2018)Alshiekh, Bloem, Ehlers, K{\"{o}}nighofer,
  Niekum, and Topcu]{rl-shielding}
M.~Alshiekh, R.~Bloem, R.~Ehlers, B.~K{\"{o}}nighofer, S.~Niekum, and U.~Topcu.
\newblock Safe reinforcement learning via shielding.
\newblock In \emph{{AAAI}}, pages 2669--2678. {AAAI} Press, 2018.

\bibitem[Ames et~al.(2019)Ames, Coogan, Egerstedt, Notomista, Sreenath, and
  Tabuada]{cbfs_overview}
A.~D. Ames, S.~Coogan, M.~Egerstedt, G.~Notomista, K.~Sreenath, and P.~Tabuada.
\newblock Control barrier functions: Theory and applications.
\newblock In \emph{2019 18th European Control Conference (ECC)}, pages
  3420--3431, 2019.
\newblock \doi{10.23919/ECC.2019.8796030}.

\bibitem[Amodei et~al.(2016)Amodei, Olah, Steinhardt, Christiano, Schulman, and
  Man{\'e}]{amodei2016}
D.~Amodei, C.~Olah, J.~Steinhardt, P.~Christiano, J.~Schulman, and D.~Man{\'e}.
\newblock Concrete problems in ai safety.
\newblock \emph{arXiv preprint arXiv:1606.06565}, 2016.

\bibitem[Arulkumaran et~al.(2017)Arulkumaran, Deisenroth, Brundage, and
  Bharath]{arulkumaran2017}
K.~Arulkumaran, M.~P. Deisenroth, M.~Brundage, and A.~A. Bharath.
\newblock A brief survey of deep reinforcement learning.
\newblock \emph{arXiv preprint arXiv:1708.05866}, 2017.

\bibitem[Balakrishnan et~al.(2022)Balakrishnan, Jak{\v{s}}i{\'c}, Aguilar,
  Ni{\v{c}}kovi{\'c}, and Deshmukh]{balakrishnan2022}
A.~Balakrishnan, S.~Jak{\v{s}}i{\'c}, E.~A. Aguilar, D.~Ni{\v{c}}kovi{\'c}, and
  J.~V. Deshmukh.
\newblock Model-free reinforcement learning for symbolic automata-encoded
  objectives.
\newblock \emph{arXiv preprint arXiv:2202.02404}, 2022.

\bibitem[Belta et~al.(2017)Belta, Yordanov, and Gol]{calin-book}
C.~Belta, B.~Yordanov, and E.~Gol.
\newblock \emph{Formal Methods for Discrete-Time Dynamical Systems}, volume~89.
\newblock Springer, 01 2017.
\newblock ISBN 978-3-319-50762-0.
\newblock \doi{10.1007/978-3-319-50763-7}.

\bibitem[Berducci et~al.(2021)Berducci, Aguilar, Ni{\v{c}}kovi{\'c}, and
  Grosu]{hierarchical-potential-rewards}
L.~Berducci, E.~A. Aguilar, D.~Ni{\v{c}}kovi{\'c}, and R.~Grosu.
\newblock Hierarchical potential-based reward shaping from task specifications.
\newblock \emph{arXiv preprint arXiv:2110.02792}, 2021.

\bibitem[Dawson et~al.(2023)Dawson, Gao, and Fan]{safe_learning_chuchu}
C.~Dawson, S.~Gao, and C.~Fan.
\newblock Safe control with learned certificates: A survey of neural lyapunov,
  barrier, and contraction methods for robotics and control.
\newblock \emph{IEEE Transactions on Robotics}, pages 1--19, 2023.
\newblock \doi{10.1109/TRO.2022.3232542}.

\bibitem[Fujimoto et~al.(2018)Fujimoto, Hoof, and Meger]{td3}
S.~Fujimoto, H.~Hoof, and D.~Meger.
\newblock Addressing function approximation error in actor-critic methods.
\newblock In \emph{International conference on machine learning}, pages
  1587--1596. PMLR, 2018.

\bibitem[Gronauer(2022)]{bullet-safety-gym}
S.~Gronauer.
\newblock Bullet-safety-gym: A framework for constrained reinforcement
  learning.
\newblock Technical report, mediaTUM, 2022.

\bibitem[Ibarz et~al.(2021)Ibarz, Tan, Finn, Kalakrishnan, Pastor, and
  Levine]{ibarz2021}
J.~Ibarz, J.~Tan, C.~Finn, M.~Kalakrishnan, P.~Pastor, and S.~Levine.
\newblock How to train your robot with deep reinforcement learning: lessons we
  have learned.
\newblock \emph{The International Journal of Robotics Research}, 40\penalty0
  (4-5):\penalty0 698--721, 2021.

\bibitem[James and Collins(2006)]{james2006analysis}
H.~W. James and E.~Collins.
\newblock An analysis of transient markov decision processes.
\newblock \emph{Journal of applied probability}, 43\penalty0 (3):\penalty0
  603--621, 2006.

\bibitem[Jothimurugan et~al.(2019)Jothimurugan, Alur, and
  Bastani]{jothimurugan2019}
K.~Jothimurugan, R.~Alur, and O.~Bastani.
\newblock A composable specification language for reinforcement learning tasks.
\newblock \emph{Advances in Neural Information Processing Systems}, 32, 2019.

\bibitem[Jothimurugan et~al.(2021)Jothimurugan, Bansal, Bastani, and
  Alur]{jothimurugan2021compositional}
K.~Jothimurugan, S.~Bansal, O.~Bastani, and R.~Alur.
\newblock Compositional reinforcement learning from logical specifications.
\newblock In A.~Beygelzimer, Y.~Dauphin, P.~Liang, and J.~W. Vaughan, editors,
  \emph{Advances in Neural Information Processing Systems}, 2021.

\bibitem[Li et~al.(2019)Li, Serlin, Yang, and Belta]{science_composition}
X.~Li, Z.~Serlin, G.~Yang, and C.~Belta.
\newblock A formal methods approach to interpretable reinforcement learning for
  robotic planning.
\newblock \emph{Science Robotics}, 4\penalty0 (37):\penalty0 eaay6276, 2019.
\newblock \doi{10.1126/scirobotics.aay6276}.

\bibitem[Nangue~Tasse et~al.(2020)Nangue~Tasse, James, and
  Rosman]{boolean-task-algebra}
G.~Nangue~Tasse, S.~D. James, and B.~Rosman.
\newblock A boolean task algebra for reinforcement learning.
\newblock In \emph{NeurIPS}, 2020.

\bibitem[Nangue~Tasse et~al.(2022{\natexlab{a}})Nangue~Tasse, James, and
  Rosman]{tasse2022}
G.~Nangue~Tasse, S.~James, and B.~Rosman.
\newblock Generalisation in lifelong reinforcement learning through logical
  composition.
\newblock In \emph{International Conference on Learning Representations},
  2022{\natexlab{a}}.

\bibitem[Nangue~Tasse et~al.(2022{\natexlab{b}})Nangue~Tasse, Jarvis, James,
  and Rosman]{skill-machines}
G.~Nangue~Tasse, D.~Jarvis, S.~James, and B.~Rosman.
\newblock Skill machines: Temporal logic composition in reinforcement learning.
\newblock \emph{CoRR}, abs/2205.12532, 2022{\natexlab{b}}.

\bibitem[Ng et~al.(1999)Ng, Harada, and Russell]{reward-shaping}
A.~Y. Ng, D.~Harada, and S.~Russell.
\newblock Policy invariance under reward transformations: Theory and
  application to reward shaping.
\newblock In \emph{{ICML}}, pages 278--287. Morgan Kaufmann, 1999.

\bibitem[Pnueli(1977)]{temporal-logic}
A.~Pnueli.
\newblock The temporal logic of programs.
\newblock In \emph{18th Annual Symposium on Foundations of Computer Science
  (sfcs 1977)}, pages 46--57, 1977.
\newblock \doi{10.1109/SFCS.1977.32}.

\bibitem[Schrittwieser et~al.(2020)Schrittwieser, Antonoglou, Hubert, Simonyan,
  Sifre, Schmitt, Guez, Lockhart, Hassabis, Graepel, et~al.]{schrittwieser2020}
J.~Schrittwieser, I.~Antonoglou, T.~Hubert, K.~Simonyan, L.~Sifre, S.~Schmitt,
  A.~Guez, E.~Lockhart, D.~Hassabis, T.~Graepel, et~al.
\newblock Mastering atari, go, chess and shogi by planning with a learned
  model.
\newblock \emph{Nature}, 588\penalty0 (7839):\penalty0 604--609, 2020.

\bibitem[Taylor and Stone(2009)]{transfer_learning}
M.~E. Taylor and P.~Stone.
\newblock Transfer learning for reinforcement learning domains: A survey.
\newblock \emph{J. Mach. Learn. Res.}, 10:\penalty0 1633–1685, dec 2009.
\newblock ISSN 1532-4435.

\bibitem[Van~Niekerk et~al.(2019)Van~Niekerk, James, Earle, and
  Rosman]{composing-value-functions}
B.~Van~Niekerk, S.~James, A.~Earle, and B.~Rosman.
\newblock Composing value functions in reinforcement learning.
\newblock In \emph{Proceedings of the 36th International Conference on Machine
  Learning}, volume~97 of \emph{Proceedings of Machine Learning Research},
  pages 6401--6409. PMLR, 09--15 Jun 2019.

\end{thebibliography}
\clearpage
\appendix
\section{Supplementary Material}

\KL{@Makai: I've started to remove comments that have been addressed. Hopefully what remains are comments that are still a bit relevant.}

\KL{I think there are some extraneous sections here, but we can double check.}




\subsection{Derivation of penalty multiplier}
In Sec.~\ref{sec:penalties}, we introduced the notion of a \emph{penalty multiplier}, \pmult, that we use to create our hierarchical reward structure. As noted in that section, this multiplier must be sufficiently large to penalize behaviors such as early termination or passing through unsafe regions. Here we derive the value of \pmult, which is used in our proofs in the following sections. We note that this value of \pmult is necessary for theoretical proofs, but in practice a smaller value can often be used to achieve the same results.

For \(g_1, g_2 \in \goal\), and proposition \(q \in \props\), let \(\mathit{AvoidPathLen}(g_1, g_2, q)\) return the length of the longest of all shortest paths between any state \(s_1 \in g_1\) and state \(s_2 \in g_2\) that passes through the minimum number of \(q\)-regions. Similarly \(\mathit{AvoidPathLen}(g_1, g_2, \neg q)\) does the same but avoiding \(\neg q\)-regions. In plain English, this is the maximum number of steps \emph{required} to get from any state in \(g_1\) to any state in \(g_2\) while passing through a minimum number of obstacle regions defined by \(q\) or \(\neg q\).

\MM{Just realized that \(C_p\) might be confusing because it may seem like it refers to \(p \in \props\) when it doesn't. We can't change that symbol because it's in the main paper, but I'm wondering if we can drop the dependence on \(p\) entirely? Maybe it should just be for avoiding all other goals?}
\KL{That makes sense to me. Although perhaps we should keep it as \pmult for now? Just to avoid confusion between \(C\) and \pmult?}
\MM{Yeah I agree, I wasn't very clear. I'm wondering if we can drop the third argument (\(p\) or \(\neg p\)) to \textit{AvoidPathLen}?} \KL{I think it's cleaner to drop the argument, but it's maybe slightly more precise to keep it. I guess if it's always a minimum-violation path we don't need it. If it's a prioritized safety path then maybe we do, since the negated regions are treated differently?}

\begin{dfn}
We call \(\pmult \in \mathbb{N}\) a \emph{penalty multiplier} when defined by:
\[
\begin{aligned}
\pmult \coloneqq  \mathit{argmin}_N & [ \forall g_1, g_2 \in \goal, q \in \props ~.~ \\
& N \geq \mathit{AvoidPathLen}(g_1, g_2, q) \wedge \\
& N \geq \mathit{AvoidPathLen}(g_1, g_2, \neg q)]
\end{aligned}
\]
\end{dfn}

Intuitively, \pmult should be set so that it is larger than the longest of all shortest paths. That way, a longer path with a lower penalty for every step of the path is preferred to a shorter path that incurs \pmult even once. This idea is used in the proofs below.

\subsection{Proof of Theorem~\ref{thm:min_viol}}


Here, we prove Theorem~\ref{thm:min_viol}. If the proposed reward hierarchy is used, then minimum-violation semantics are produced.

We introduce the following notation for the purposes of this proof and the proofs that follow:
\begin{itemize}
    \item Variables that capture path length:
    \begin{itemize}
    \item \(l_{unlabeled}\) - the total length of states in a path that do not produce a symbol
    \item \(l_{badLabel}\) - the total length of states in a path that produce an undesired label
    \item \(l_{worstPassThrough}\) - the total length of states in a path that produce a negated label
    \end{itemize}
    \item Indicator functions for termination:
    \begin{itemize}
    \item \(1_{goal}\) - 1 if terminates at goal, 0 otherwise
    \item \(1_{badTerm}\) - 1 if terminates at bad state, 0 otherwise
    \item \(1_{worstTerm}\) - 1 if terminates at negated state, 0 otherwise
    \end{itemize}
\end{itemize}
Note that the indicator functions are mutually exclusive. That is, only one of them can equal \(1\). Further, let \(l_{max} \coloneqq  l_{unlabeled} + l_{badLabel} + l_{worstPassThrough}\). This is the path length and by definition \(\pmult \geq l_{max}\) for any \emph{optimal} path.

The total undiscounted reward received for a path between two states, \(s_1\) and \(s_2\) is the following:
\begin{multline}
    R = R_{goal}*1_{goal}\\+R_{step}\left(\pmult^2*1_{badTerm}+\pmult^3*1_{worstTerm}+\pmult^4*1_{neverTerm}\right)\\ + R_{step}\left(l_{unlabeled}+\pmult*l_{badLabel}+ \pmult^2*l_{worstPassThrough}\right)\;,
\end{multline}
where the first line is the reward for terminating at the goal, the second line consists of penalties for terminating elsewhere, and the third line consists of penalties for passing through other regions.

Since \(R_{step}<0\), the first line is always greater than the second two lines for any path, since it is always non-negative. 

We must prove two properties:
\begin{enumerate}
    \item If a pure path exists, it will be taken, otherwise a minimally violating path will be taken; and
    \item If a goal is reachable by \emph{any} path, the minimum-violation path will be taken, instead of terminating early at an undesired goal.
\end{enumerate}

If a pure 
path exists, then \(1_{goal}\) is achievable and there exists a path consisting entirely of \(l_{unlabeled}\). Since \(l_{max}\geq l_{unlabeled}\) and \(R_{step}\leq 0\), we know that
\begin{align}
    R_{pure} &= R_{goal} + R_{step}l_{unlabeled}\\
     &\geq R_{goal} + R_{step}l_{max}\\
     &\geq R_{goal} + R_{step}\pmult\:.
\end{align}
The last line is equivalent to the total reward for a path of length one, consisting only of a bad label. This implies that a pure path always returns a higher reward than a path with even a single bad label. 
This is because \pmult exceeds \(l_{max}\). Therefore, the same logic holds for states that pass through negated goals or achieve other pass-through penalties, due to the increase of each penalty by \pmult.

\begin{remark}
Note that we assume \(\pmult\geq l_{max}\). While this is necessary in theory, in practice it is typically sufficient to use a penalty of relatively large magnitude. Relaxing this assumption in practice is useful for speeding up convergence.
\end{remark}

In the worst case, the minimum-violation path reaches goal \(g\) while taking a path that achieves a penalty for a bad step at every step. The total accumulated reward for such a path is
\begin{align}
    R_{worstMinViol} &=  R_{goal} + R_{step}\pmult l_{badLabel}\label{eq:minViol1}\\
    & \geq R_{goal} + R_{step}\pmult l_{max}\label{eq:minViol2}\\
    & \geq R_{goal} + R_{step}\pmult^2\label{eq:minViol3}\:,
\end{align}
while the reward for a path of length 1 that terminates at any other goal is
\begin{align}
    R_{earlyTerm} &= R_{step}\pmult^2+R_{step}\\
    &\leq R_{step}\pmult^2+R_{goal}\\
    &\leq R_{worstMinViol}\:,
\end{align}
where the first inequality follows from the fact that \(R_{goal}>R_{step}\), and the second inequality follows from~\eqref{eq:minViol1}-\eqref{eq:minViol3}. Since a path of length 1 that terminates at any other goal is worse than the worst-case minimum-violation path, any longer path that terminates at any other goal is also worse than the longest minimum-violation path.

\subsection{Proof of Theorem~\ref{thm:priority}}

For negated tasks trained with~\eqref{eq:negation}, we can follow similar logic. Here, a path of length one that passes through a state containing a symbol that violates prioritized safety produces a reward of
\begin{align}
    R_{priorSafety} &= R_{goal} + R_{step}\pmult^2\\
&\geq R_{goal} + R_{step}\pmult l_{max}\\
&= R_{goal} + R_{badStep}l_{max}\:,
\end{align}
where the second line follows from the fact that \(\pmult\geq l_{max}\). The last line is equivalent to the reward accrued by a path of length \(l_{max}-1\) that exclusively passes through states that give a reward of \(R_{badStep}\). Therefore, any path that passes through a negated label achieves a reward that is worse than the longest possible path that passes through any other (non-goal) labels. Otherwise, the hierarchy follows the same pattern as minimum-violation policies.

\KL{I suspect early termination theoretically requires an additional penalty on terminating in an undesired state. I see a couple of options. 1) include that reward in the text, even if we don't use it; 2) don't include it and get hand-wavy about the proof (i.e., the proof follows exactly from minimum violation). thoughts?} \MM{Undesired here means with no label? Because we do have a bad termination for states that are not goal. I think we can include it in the text even if we don't use it for theoretical completeness. I guess the other way to solve that is ensure that the goal reward is large enough.} \KL{I think the issue is that a non-negated, non-goal labeled state gets an intermediate penalty that is equivalent to the worst step penalty. So all of a sudden, early termination in such a state is not as bad.}

\subsection{Proof of Theorems~\ref{thm:min_viol_comp} and~\ref{thm:priority_comp}}

For composition to work, we remind the reader of the following assumptions:
\begin{itemize}
    \item The policy, extended Q-value functions, and/or value functions have converged to their optimal values (Assumption~\ref{ass:policy}); and
    \item Rewards and penalties only differ on labeled states, defined by Defs.~\eqref{eq:min_viol_reward} and~\eqref{eq:negation}, as applicable.
\end{itemize}

\KL{Changed task from \(T\) to \(\mathcal{M}\) to be consistent with main body of the paper.}
For an extended Q-value function, each tuple \((\mathcal{M},g,s)\) of task, goal, and state contains information about the value of each possible action and therefore the best action to take from state \(s\). Taken in sequence, this induces a minimum-violation path for that tuple, with respect to the task-goal pair.

Previous sections in this work demonstrated that a given policy is optimal in this sense with respect to an arbitrary combination of tasks, goals, and bad states. That is, the reward structure will be respected by any policy obtained under that structure. What remains to be proven is that the composition of such policies continues to respect that structure.

\begin{remark}
    Note that the proof of composition for prioritized safety semantics only holds for negation in the case that there is path free of bad states (e.g. states satisfying \(\phi_B\) from Definition \ref{def:priority}). If no such path exists, it is possible to obtain chattering in the policy even under Assumption \ref{ass:prioritized_safety}.
\end{remark}

For the following proofs, we introduce Lemma~\ref{lem:path_enc} below.

\begin{lemma}\label{lem:path_enc}
    For a given policy \(\bar{\pi}\) trained under the reward structure given by~\eqref{eq:min_viol_reward} (resp.~\eqref{eq:negation}), the policy from any given state encodes a path that corresponds to the shortest minimum-violation (resp. prioritized safety) path from that state.
\end{lemma}

\begin{proofsketch}
    We assume a deterministic MDP. Then, using the definitions of optimal policies, this lemma follows directly from Theorems~\ref{thm:min_viol} \&~\ref{thm:priority}, respectively.
\end{proofsketch}


\subsubsection{Disjunction}
For two extended Q-value functions, \(\Q_1\) and \(\Q_2\), the disjunction of the two is defined as
\begin{equation}
    \Q_{1\vee 2}(s,g,a) = \max\{\Q_1(s,g,a),\Q_2(s,g,a)\}\;,
\end{equation}
and the resulting policy is
\begin{equation}
    \bar{\pi}_{1\vee 2}(s) = \arg\max_{a\in\mathcal{A}}\{\max_{g\in\goal}\max\{\Q_1(s,g,a),\Q_2(s,g,a)\}\}\:.
\end{equation}

We wish to prove the following:
\begin{itemize}
    \item For non-negated policies, and negated policies with pure paths, there is no chattering (based on path calculus)
    \item For non-negated policies, and negated policies with pure paths, the optimal policy satisfies one or the other extended Q-value functions (i.e., satisfies \(1 \vee 2\). 
    )
\end{itemize}


\begin{proof}\label{proof:discrete-disj}
For policies \(\bar{\pi}_1\) and \(\bar{\pi}_2\) that each encode minimum-violation semantics and an arbitrary state \(s\), we can assume without loss of generality that \(\bar{\pi}_1\) encodes the shorter minimum-violation path to some goal \(g\). Note that \(\bar{\pi}_2\) may encode a path for either the same goal \(g\) or a different goal \(g'\). \KL{Added to clarify about potentially different goals.} If we call their corresponding executions \(\exec_1\) and \(\exec_2\), then we know (by definition) that \(\vert{\upharpoonright}_L^+(\exec_1)\vert\leq\vert{\upharpoonright}_L^+(\exec_2)\vert\), so the sequence of labeled states must be shorter for \(\exec_1\) as well. Since they are both minimum-violation, and all rewards other than the goal reward are negative, the shorter path has a higher cumulative reward-to-go, and therefore \(\bar{V}_1(s)\geq\bar{V}_2(s)\). Because \(\bar{\pi}\) encodes an optimal path for at least one goal \(g\), the same reasoning can be applied for \(s'\), the state reached from \(s\) by applying \(\bar{\pi}(s)\). \KL{Added this line for ``induction" purposes.} This is equivalent to the definition of disjunction, and therefore respects minimum-violation semantics.
\end{proof}

\subsubsection{Conjunction}
For two extended Q-value functions, \(\Q_1\) and \(\Q_2\), the conjunction of the two is defined as
\begin{equation}
    \Q_{1\wedge 2}(s,g,a) = \min\{\Q1(s,g,a),\Q_2(s,g,a)\}\;,
\end{equation}
and the resulting policy is
\begin{equation}
    \bar{\pi}_{1\wedge 2}(s) = \arg\max_{a\in\mathcal{A}}\{\max_{g\in\goal}\min\{\Q_1(s,g,a),\Q_2(s,g,a)\}\}\:.
\end{equation}


\begin{assumption}
For this proof, we assume that conjunction is semantically meaningful. I.e., for \(\mathcal{M}_1\wedge\mathcal{M}_2\), we assume there exists at least one goal that satisfies the conjunction of \(\mathcal{M}_1\) and \(\mathcal{M}_2\).
\end{assumption}


\begin{proof}\label{proof:discrete-conj}
    Consider two tasks \(\mathcal{M}_{p_1}\) and \(\mathcal{M}_{p_2}\), and their corresponding optimal extended value functions, \(\Q^*_1\) and \(\Q^*_2\). Let \(g^* \in \goal\) be the goal that satisfies both tasks with the lowest-penalty path from \(s\). Furthermore, let \(g \in \goal\) be a goal that satisfies both tasks but has no lower penalty path from \(s\) than \(g^*\), and \(g^\prime \in \goal\) be a goal that does not satisfy at least one of the tasks.

    For simplicity, we first consider minimum-violation semantics. By construction, \(\mathit{max}_{a \in \mathcal{A}}\Q^*_1(s, g^*, a) = \mathit{max}_{a \in \mathcal{A}}\Q^*_2(s, g^*, a)\) because \(g^*\) satisfies both tasks and the penalties are identical in minimum violation. Thus, the maximizing action is identical for both value functions. There can be more than one maximizing action, in which case we have no preference.

    Furthermore, we know that \(\mathit{max}_{g\in\goal, a\in\mathcal{A}}\Q^*_i(s, g, a) < \mathit{max}_{g\in\goal, a\in\mathcal{A}}\Q^*_i(s, g^*, a)\) because there is no lower penalty path to \(g\). Applying the minimum operator for conjunction preserves this element-wise inequality and thus, the composed policy will choose actions for the path to \(g^*\).

    Finally, because the \(\pmult^2 R_{step}\) penalty on bad terminations, either \(\Q^*_1(s, g^\prime, a)\) or \(\Q^*_2(s, g^\prime, a)\) has a low value compared to the paths to \(g^*\).
    Due to the minimum operation in conjunction, this is the preserved value for \(\Q^*_{1\wedge 2}\) and \( \mathit{max}_{g^*\in\goal,a\in\mathcal{A}}\Q^*_{1\wedge 2}(s, g, a) < \mathit{max}_{g^*\in\goal,a\in\mathcal{A}}\Q^*_{1\wedge 2}(s, g^*, a)\). Thus, the composed policy will choose actions corresponding to paths to \(g^*\).

    For prioritized safety semantics, we rely on Assumption \ref{ass:prioritized_safety}. Using Option \ref{it:ps-opt1} from the assumption, we have only a single conjuncted safety constraint. Unsafe actions leading to \(R_{worststep}\) penalties have low values that dominate the minimum operation. The termination penalties are the same. The chosen path will correspond to the best prioritized safety path to a goal that satisfies both the safety constraint and the other conjunct. See Section \ref{sec:sup-prior-safety} for more details and an explanation of Option \ref{it:ps-opt2}.
\end{proof}

\subsection{Proof of Theorem~\ref{thm:continuous}}
To extend Boolean task composition to continuous action spaces for minimum-violation semantics, we train negated policies separately. We now prove that disjunction and conjunction over these positive and negated learned tasks behaves as expected. 


\subsubsection{Disjunction}
\MM{Use \(\Q\)}
For disjunction of two tasks, the associated \(\Q\)-function is~\cite{boolean-task-algebra}:
\begin{equation}
    \Q^*_1 \vee \Q^*_2(s,g,a) := \max\lbrace \Q_1^*(s,g,a), \Q_2^*(s,g,a)\rbrace]\:.
\end{equation}
For discrete action spaces, the \(\max\) can be determined by comparing over all actions in \(\mathcal{A}\). For continuous action spaces, that is not feasible, so we must determine the optimal choice analytically.

From a policy standpoint, let's consider \(\Q^*_1 \vee \Q^*_2\). The optimal action will be \(\pi^*_{\Q^*_1 \vee \Q^*_2}:S\rightarrow \mathcal{A}\) and is given by 
\KL{I think we need a max over \(g\) as well? Likewise, it needs to carry through the rest of these proofs.} \MM{Or, we just include \(g\) in the result and make a comment about maxing over g.}
\begin{equation}\label{eq:q_or}
\begin{aligned}
    \pi^*_{Q^*_1 \vee Q^*_2}(s) &= \argmax_{a \in \mathcal{A}}[\max_{g\in\goal} {\Q^*_1 \vee \Q^*_2}(s, g, a)] \\
    \text{where} &\\
    {\Q^*_1 \vee \Q^*_2}(s, g, a) &= \max\lbrace \Q_1^*(s,g,a), \Q_2^*(s,g,a)\rbrace\:.
\end{aligned}
\end{equation}
Note that we can swap the order of maximizing over \(a\in\mathcal{A}\) and \(g\in\goal\). Thus, we denote a policy \(\pi(s,g)\) as the optimal action for reaching goal \(g\) with the minimum number of penalties. Since we know by definition that \(\Q^*(s,g,\cdot)\) is maximized by the corresponding \(\pi^*(s, g)\), then the solution to~\eqref{eq:q_or} is either \(\pi^*_1(s)\) or \(\pi^*_2(s)\)
where \(\pi^*_i(s) \coloneqq \mathit{max}_{g \in \goal}\pi^*_i(s, g)\). 
That is, the global maximum for \(\Q^*_1\vee \Q^*_2(s,g,\cdot)\) is either the global maximum of \(\Q^*_1(s,g,\cdot)\) or the global maximum of \(\Q^*_2(s,g,\cdot)\).

Therefore, the optimal policy for disjunction of two tasks is
\begin{equation}
    \pi^*_{\Q^*_1 \vee \Q^*_2}(s, g) = 
    \begin{cases}
      \pi^*_1(s, g) & \text{if \(\Q^*_1(s,g,\pi_1^*(s, g))\geq \Q^*_2(s,g,\pi_2^*(s, g))\)}\\
      \pi^*_2(s, g) &  \text{otherwise}\:.
    \end{cases}  
\end{equation}

The optimal action is chosen using \(\mathit{max}_{g\in\goal} \pi^*_{\Q^*_1 \vee \Q^*_2}(s, g)\).
\begin{proofsketch}
The proof of correctness follows the same reasoning as Proof \ref{proof:discrete-disj}.
\end{proofsketch}

\subsubsection{Conjunction}
\MM{Need to update this as well}
We follow similar logic to reason about conjunction. For conjunction of two tasks, the associated \(\Q\)-function is~\cite{boolean-task-algebra}:
\begin{equation}
    \Q^*_1 \wedge \Q^*_2(s,g,a) := \min\lbrace \Q_1^*(s,g,a), \Q_2^*(s,g,a)\rbrace]\:.
\end{equation} The best action for a conjunction of terms is written
\begin{equation}\label{eq:q_conj}
    \pi^*_{Q^*_1\wedge Q^*_2}(s)=\argmax_{a\in \mathcal{A}}[\max_{g\in\goal}\min\lbrace \Q_1^*(s,g,a), \Q_2^*(s,g,a)\rbrace]\:.
\end{equation}

The resulting policy is therefore
\begin{equation}
    \pi^*_{\Q^*_1 \wedge \Q^*_2}(s, g) = 
    \begin{cases}
      \pi^*_1(s, g) & \text{if \(\Q^*_1(s,g,\pi_1^*(s, g))\leq \Q^*_2(s,g,\pi_2^*(s, g))\)}\\
      \pi^*_2(s, g) &  \text{otherwise}\:.
    \end{cases}  
\end{equation}

\begin{proofsketch}
For minimum-violation semantics or prioritized safety semantics with Assumption \ref{ass:prioritized_safety}, we are guaranteed that the composed policies either agree on the value of a path for the same satisfying goal with a high value, or is dominated by a low-value (due to a non-satisfying goal, unsafe action, or long path) due to the minimum operation. Note that there could be more than one optimal path and thus different possible actions. However, for goals that satisfy both tasks, they must agree on the value and thus we do not have a preference between these two paths. Thus, the optimal action is encoded by either \(\pi^*_1\)  or \(\pi^*_2\). The justification follows the same reasoning as Proof \ref{proof:discrete-conj}.
\end{proofsketch}

\MM{Is this a reasonable clarification of our claim about prioritized safety in continuous action spaces? Last paragraph of Section \ref{sec:continuous}.}

Note, we face the same limitations for prioritized safety in continuous action spaces as in discrete action spaces. However, we can rely on the same assumptions in Assumption \ref{ass:prioritized_safety} and leverage the same approaches used in Section \ref{sec:sup-prior-safety} to lift this notion to continuous action spaces as well.

\subsection{Composition as union and intersection}
In this penalty-based reward structure, conjunction (\(\wedge\)) and disjunction (\(\vee\)) have intuitive semantics over tasks \(\mathcal{M}_{p_1}\) and \(\mathcal{M}_{p_2}\):
\begin{itemize}
    \item conjunction takes the intersection (\(\cap\)) of task \(p_1\)-regions and \(p_2\)-regions to be reached, and the union (\(\cup\)) of \(\neg p_1\)-regions and \(\neg p_2\)-regions to be avoided
    \item disjunction takes the union (\(\cup\)) of task \(p_1\)-regions and \(p_2\)-regions to be reached, and the intersection (\(\cap\)) of \(\neg p_1\)-regions and \(\neg p_2\)-regions to be avoided
\end{itemize}

\section{Experimental Details and Analysis}
\begin{figure}
     \centering
     \begin{subfigure}[b]{0.3\textwidth}
         \centering
         \includegraphics[width=\textwidth]{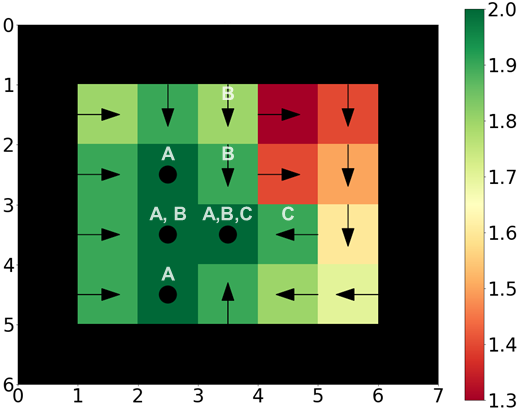}
         \caption{\(A\)}
     \end{subfigure}
     \hfill
     \begin{subfigure}[b]{0.3\textwidth}
         \centering
         \includegraphics[width=\textwidth]{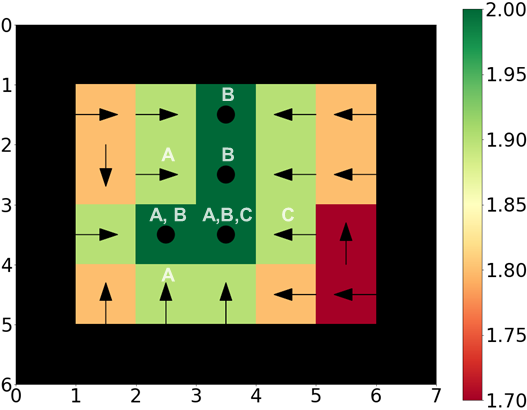}
         \caption{\(B\)}
     \end{subfigure}
     \hfill
     \begin{subfigure}[b]{0.3\textwidth}
         \centering   
         \includegraphics[width=\textwidth]{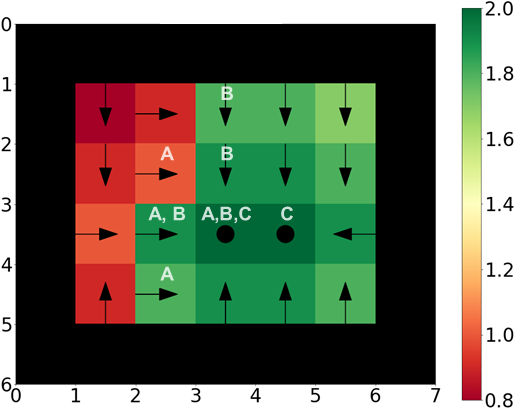}
         \caption{\(C\)}
     \end{subfigure}
     \\
     \begin{subfigure}[b]{0.3\textwidth}
         \centering
         \includegraphics[width=\textwidth]{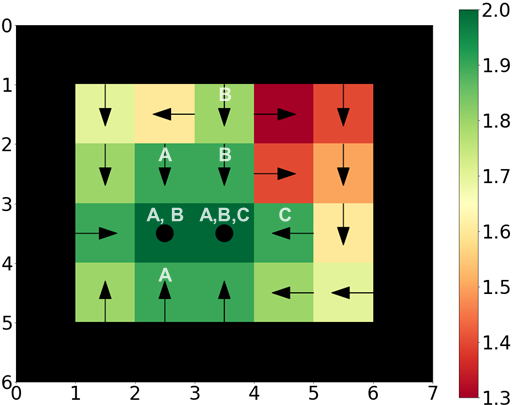}
         \caption{\(A \wedge B\)}
     \end{subfigure}
     \hfill
     \begin{subfigure}[b]{0.3\textwidth}
         \centering
         \includegraphics[width=\textwidth]{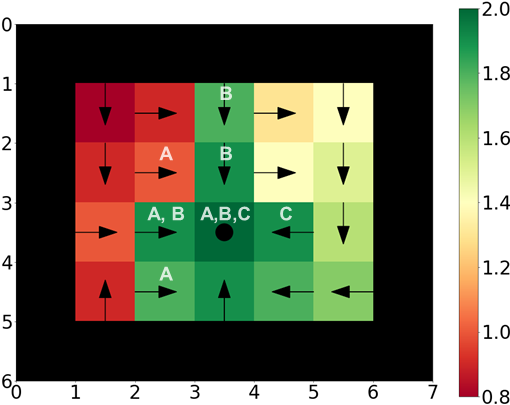}
         \caption{\(B \wedge C\)}
     \end{subfigure}
     \hfill
     \begin{subfigure}[b]{0.3\textwidth}
         \centering   
         \includegraphics[width=\textwidth]{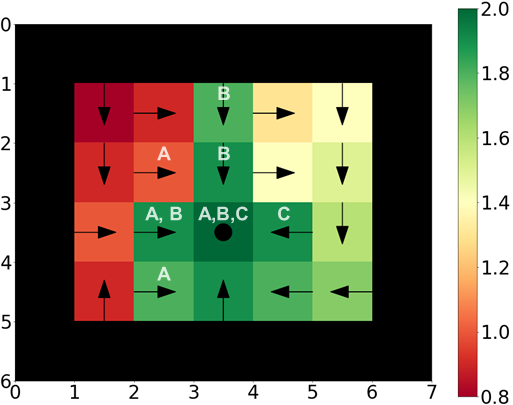}
         \caption{\(A \wedge B \wedge C\)}
     \end{subfigure}
     \\
     \begin{subfigure}[b]{0.3\textwidth}
         \centering
         \includegraphics[width=\textwidth]{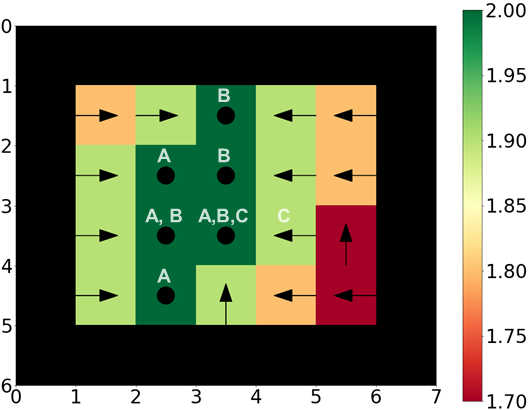}
         \caption{\(A \vee B\)}
     \end{subfigure}
     \hfill
     \begin{subfigure}[b]{0.3\textwidth}
         \centering
         \includegraphics[width=\textwidth]{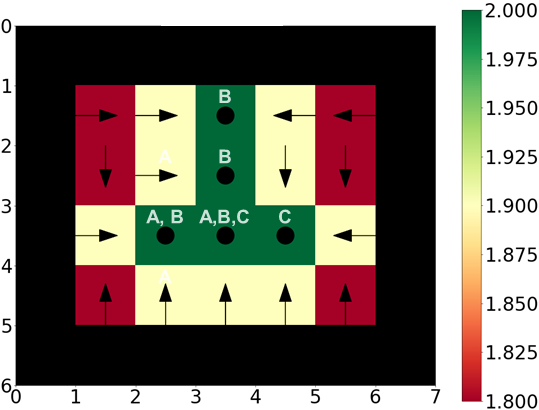}
         \caption{\(B \vee C\)}
     \end{subfigure}
     \hfill
     \begin{subfigure}[b]{0.3\textwidth}
         \centering   
         \includegraphics[width=\textwidth]{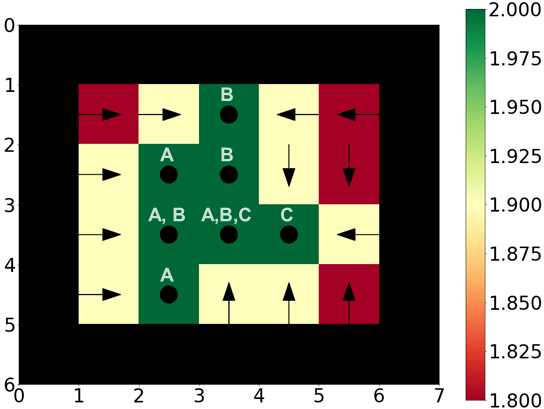}
         \caption{\(A \vee B \vee C\)}
     \end{subfigure}
     \\
     \begin{subfigure}[b]{0.3\textwidth}
         \centering
         \includegraphics[width=\textwidth]{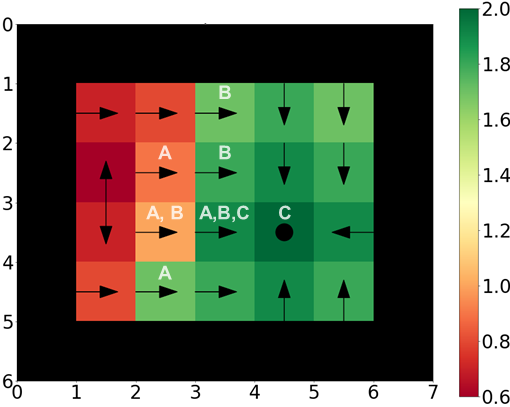}
         \caption{\(\text{not-}A \wedge \text{not-}B\)}
     \end{subfigure}
     \hfill
     \begin{subfigure}[b]{0.3\textwidth}
         \centering
         \includegraphics[width=\textwidth]{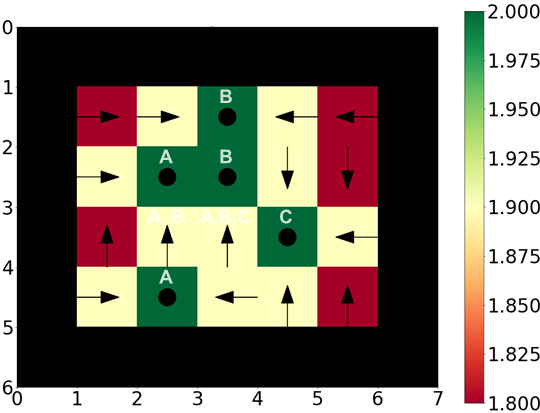}
         \caption{\(\text{not-}A \vee \text{not-}B\)}
     \end{subfigure}
     \hfill
     \begin{subfigure}[b]{0.3\textwidth}
         \centering   
         \includegraphics[width=\textwidth]{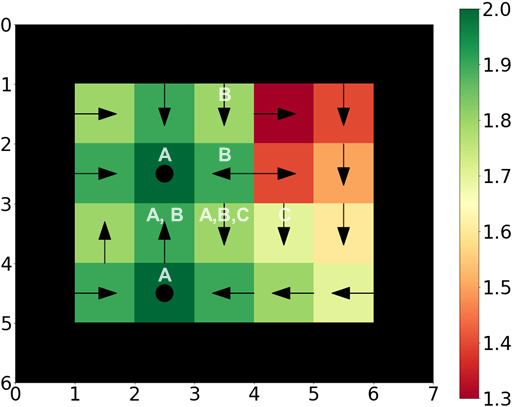}
         \caption{\(A \wedge \text{not-}B \wedge \text{not-}C\)}
     \end{subfigure}
     \caption{Several task composition examples in the example environment using value iteration. \KL{Maybe include the fact that negation is prioritized safety?}\label{fig:sup-value-iteration}}
\end{figure}

In this section we expand on the experimental results of Sec. \ref{sec:experiments}. 
For Environment \ref{env:grid}, we used our implementation of value iteration for extended Q-value functions. \KL{Does it matter that it was handwritten? Is this just to point out that it wasn't from some library?} \MM{Yeah, that's right. I'll call it something else, maybe ``our implementation"} Fig. \ref{fig:sup-value-iteration} depicts several other example compositions in the simple grid world. Recall that these policies are optimal because this is converged value iteration and has no function approximation. The negated policies in this figure are learned for prioritized safety (as opposed to the minimum violation semantics given by the negation operator in \eqref{eq:neg_op}). We did not utilize Assumption \ref{ass:prioritized_safety} and the associated approaches detailed in \ref{sec:sup-prior-safety}. This demonstrates that prioritized safety often works without chattering in practice, despite requiring additional assumptions for a formal guarantee.

\begin{figure}
     \centering
     \begin{subfigure}[b]{0.3\textwidth}
         \centering
         \includegraphics[width=\textwidth]{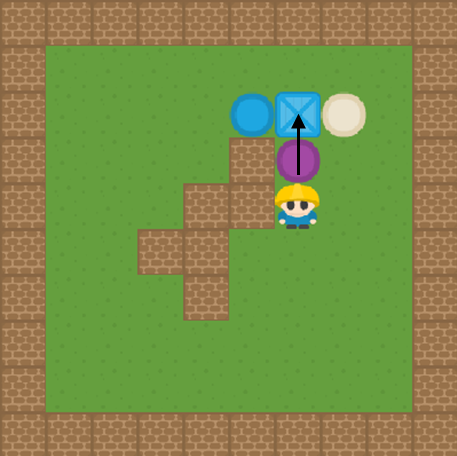}
         \caption{\(\mathit{square}\)}
     \end{subfigure}
     \hfill
     \begin{subfigure}[b]{0.3\textwidth}
         \centering
         \includegraphics[width=\textwidth]{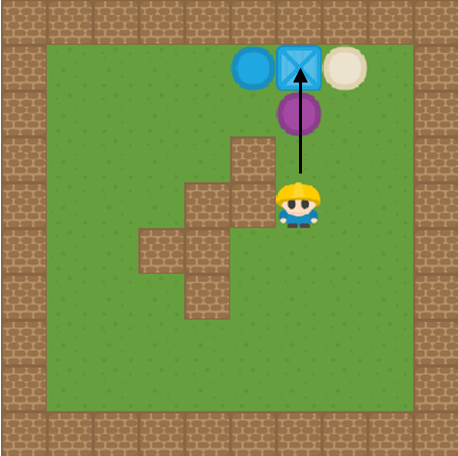}
         \caption{\(\mathit{blue} \wedge \neg \mathit{circle}\)}
     \end{subfigure}
     \hfill
     \begin{subfigure}[b]{0.3\textwidth}
         \centering   
         \includegraphics[width=\textwidth]{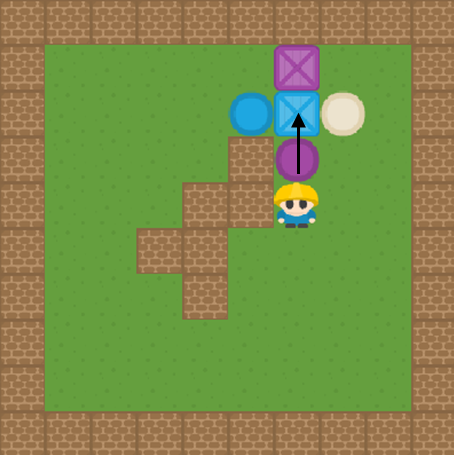}
         \caption{\(\mathit{square} \wedge \neg \mathit{purple}\)}
     \end{subfigure}
     \caption{This figures shows the semantics of the original (penalty-free) Boolean Task Algebra on the same examples used in Fig. \ref{fig:discrete_pure_square}-\ref{fig:discrete_prioritized_safety}.\label{fig:orig-discrete-semantics}}
\end{figure}

For Environment \ref{env:repoman}, we use a modified version of the code from \citet{boolean-task-algebra}. We made updates to include penalties and perform curriculum learning in the following order:
\begin{enumerate}
\item train a model to reach uniformly sampled goal items without any penalties (or use a pre-trained model from \citet{boolean-task-algebra})
\begin{itemize}
\item allocated 2M steps, but converges earlier (~1M steps)
\end{itemize}
\item refine the model by further training on random environments but with penalties required for safety added
\begin{itemize}
\item allocated 2M steps, but also converges earlier (<1M steps)
\end{itemize}
\item refine the model further on closely spaced items
\begin{itemize}
\item 20-60K steps
\item this is because the model learns to perform well on random environments, but those tend to have spread out items rather than tightly constrained layouts
\end{itemize}
\end{enumerate}
Fig. \ref{fig:orig-discrete-semantics} shows the expected semantics of the original (penalty-free) Boolean Task Algebra introduced in \citet{boolean-task-algebra}.

For Environment \ref{env:bullet-safety} we use a modified version of the TD3 code from \citet{td3}. Unlike the previous two environments, we train a dedicated policy for each goal, which we found to perform better in this case. We trained on random environments for up to 4M steps. To achieve the policies used in the demo, we trained on that static environment for an additional 1M steps. We incorporated penalties immediately for this environment, because there did not appear to be any advantage to curriculum learning in this case.

\section{Approximated Training Steps to Achieve All Tasks}

\begin{figure}
	\centering
	\begin{subfigure}{\textwidth}
		\includegraphics[width=\textwidth]{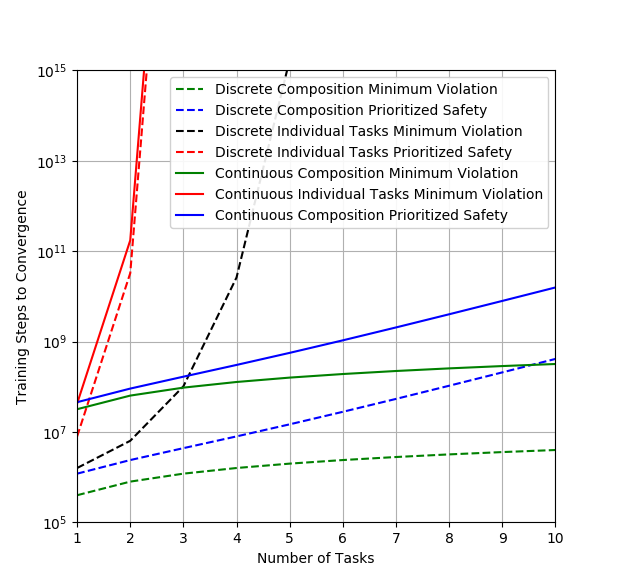}
	\end{subfigure}
 \caption{Comparison of approximated training steps required for each method}
 \label{fig:task-steps}
\end{figure}

\MM{@Zack please fix anything here and/or add more details. I can provide more info on the minimum violation / prioritized safety difference, but if you could just ensure the description of the plots is accurate, that would be great!} 
Fig.~\ref{fig:task-steps} shows a comparison of approximate training steps required for different approaches to achieve all Boolean combinations of an increasing number of tasks. We considered a policy converged when a rolling average of evaluated rewards settles within 3\% of the overall maximum reward during training in the discrete and continuous environments. We approximately extrapolate the training steps for each approach by multiplying the number of policies that need to be learned for different approaches by the time to train one task or the combination of base tasks. We compare the (approximated) total times to achieve all tasks for:
\begin{inparaenum}[i)]
\item learning extended value functions for positive tasks only (using analytic negation), for composition with minimum violation;
\item learning extended value functions for positive tasks and all possible safety constraints (Option \ref{it:ps-opt1} of Assumption \ref{ass:prioritized_safety}), for composition with prioritized safety;
\item learning all Boolean combinations (regular value functions) directly for positive tasks only, which we call individual tasks with minimum violation; and
\item learning all Boolean combinations (regular value functions) directly for positive and negated tasks, which we call individual tasks with prioritized safety.
\end{inparaenum}
This plot is on a log scale, but the difference is still exponential because the number of possible compositions is doubly exponential in the individual task combination traning cases. We observe that the cost of learning negated tasks for prioritized safety is negligible at this scale. Furthermore, even learning all possible safety constraints scales significantly better than learning all individual tasks. We also note that many use cases do not require learning all possible safety constraints. The environment of interest may only have a subset of the space of safety constraints that is relevant. Also, note that moving from the discrete to continuous domain, there is an increase in training time within a given approach, but that overall increase in training time is negligible in comparison to the individual learning cases as the number of tasks considered grows. 

\section{Prioritized safety assumption and solutions}\label{sec:sup-prior-safety}
Here we provide more information on Assumption \ref{ass:prioritized_safety} of Section \ref{sec:prior_safety}. Recall that prioritized safety puts extra weight on avoiding specifically negated region labels and that we train negated tasks explicitly. Due to the avoidance assymmetry, we cannot guarantee that composition between negated tasks works as desired in every case (despite it often working practice) without an additional assumption. The primary failure mode is chattering (infinite loops). 

Chattering occurs when the optimal policies for negated tasks have not encoded the same pure paths and the environment layout causes them to disagree. In this case, the best action in the composed policy might not be the optimal action from either negated task. Note that only optimal paths are stored in an extended value function. Any non-optimal action typically reflects the value of moving off and back onto an optimal path.

\begin{figure}
    \centering
    \resizebox{0.5\linewidth}{!}{
    \begin{tikzpicture}
    \draw[step=1.0cm,color=gray] (0,0) grid (5,4);
    \draw[step=1.0cm,color=black,line width=.5mm] (1,0) grid (2,1);
    \draw[step=1.0cm,color=black,line width=.5mm] (1,1) grid (2,2);
    \draw[step=1.0cm,color=black,line width=.5mm] (1,2) grid (2,3);
    \draw[step=1.0cm,color=black,line width=.5mm] (2,1) grid (3,2);
    \draw[color=black,line width=.5mm] (2,2) rectangle (3,4);
    \draw[step=1.0cm,color=black,line width=.5mm] (3,1) grid (4,2);
    
    \node[font=\small,fill=white,opacity=.7] at (1.5,2.5) {\(A\)};
    \node[font=\small,fill=white,opacity=.7] at (1.5,1.5) {\(A,B\)};
    \node[font=\small,fill=white,opacity=.7] at (1.5,0.5) {\(A\)};
    \node[font=\small,fill=white,opacity=.7] at (2.5,3.5) {\(B\)};
    \node[font=\small,fill=white,opacity=.7] at (2.5,2.5) {\(B\)};
    \node[font=\small,fill=white,opacity=.7] at (2.5,1.5) {\(A,B,C\)};
    \node[font=\small,fill=white,opacity=.7] at (3.5,1.5) {\(C\)};
    
    \draw (3,4) node[anchor=north west, font=\scriptsize] {\(\bigstar\)};
    
    \draw [->,red,line width=.5mm](3.5 , 3.85) -- (4.5, 3.85);
    \draw [->,red,line width=.5mm](4.5 , 3.7) -- (3.5, 3.7);
    
    \draw [->,blue,line width=.5mm](1.5 , 3.3) -- (2.5, 3.3);
    \draw [->,blue,line width=.5mm](2.5 , 3.3) -- (3.6, 3.3);
    \draw [->,blue,line width=.5mm](3.6 , 3.3) -- (3.6, 2.2);
    \draw [->,blue,line width=.5mm](3.6 , 2.2) -- (3.6, 1.6);
    \end{tikzpicture}}\caption{Example optimal vs non-optimal path\label{fig:optimal-path}}
\end{figure}
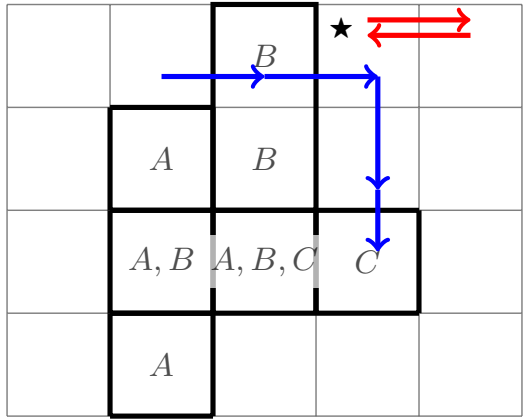

For example, consider the optimal path depicted in blue in Figure \ref{fig:optimal-path}. Let the value for action \(a\) and goal \(C\) at the \(\bigstar\) state be denoted \(\Q^*(\bigstar, C, a)\). Then \(\Q^*(\bigstar, C, \mathit{right}) = \Q^*(\bigstar, C, \mathit{down}) + 2R_{step}\). This is because stepping to the right does not fall on an optimal path and thus the value function is encoding the steps depicted by the red arrows that step away and back onto the optimal path. Depending on the region layout, this can lead to chattering under prioritized safety semantics. For the remainder of this section, we focus on conjunctions in NNF because this is how chattering arises.
\KL{This all makes sense to me. But we've been in the weeds on this issue a few times. Still, I don't think there's a clearer way to explain it.}

Assumption \ref{ass:prioritized_safety} proposed two options for circumventing this challenge:
\begin{enumerate}
\item only a single negated policy is used in composition at a time during deployment [note this policy can be more complex than a single task (e.g., learn \(\mathcal{M}_{\neg p_1 \wedge \neg p_2}\)), but it must be learned ahead of time]; or\label{it:ps-opt1},
\item any prioritized safety path in the environment of interest will only have to pass through a known, finite number, \(k\), of non-satisfying regions in \(\goal\) and we train an extended value function that maintains more corresponding paths.\label{it:ps-opt2}
\end{enumerate}

Option \ref{it:ps-opt1} can still represent any Boolean formula, but any safety constraints that may be requested at deployment must be known and trained in advance. This option makes sense in scenarios where potential safety constraints are clear at training time and they can be pre-trained. Note that any safety constraint that must be followed in all cases can trivially be included in the training procedure for every task. Thus, we are concerned with safety constraints that need to be enabled or disabled at deployment time based on user input.

Option \ref{it:ps-opt2} can be composed arbitrarily at deployment but requires learning additional Q-table entries, each of which has a lower penalty for passing through different subsets of \(\goal\).
Let a \emph{safety extended value function} \(\bar{\Q}(s, g, G_{ok}, a): \mathcal{S}\times\mathcal{G}\times2^\mathcal{G}\times\mathcal{A}\rightarrow\mathbb{R}\) be an extended Q-value function that behaves identically to an extended Q-value function except that it provides a lighter penalty for passing through any \(g_{ok} \in G_{ok}\). This is accomplished by shifting rewards as needed via the \(\pmult\) multiplier. 
\KL{I like this. This looks like a more generalized form of what we put together. We could consider this for a more unified presentation of the material in the future.} \MM{Cool, good idea!}

A safety extended value function maintains additional optimal paths for scenarios in which passing through certain goal regions is allowed. Thus, it maintains more paths. The rewards are structured to still prefer pure paths. The assumption in Option \ref{it:ps-opt2} states that all required members of \(G_{ok}\) are known at training time. This depends on domain knowledge about the environment. It might be that the environment regions are not closely packed and there are only a few scenarios in which passing through a different goal region for prioritized safety is required. Those goal regions can be added to \(G_{ok}\) to maintain paths that are allowed to pass through those regions at a smaller penalty. With these additional paths, Boolean composition works the same way. Paths that satisfy all the composed tasks have the highest value because they agree. If there is no pure path, then one of the paths passing through an allowed label region may be the best path. Negated tasks will dominate saved paths that pass through violating regions with a very negative value, keeping them from being chosen as the optimal action.

This approach increases the number of required Q-table entries by a factor of \(k\) and if \(k = 2^{\vert \goal \vert}\) then it can handle any possible prioritized minimum-violation path for any placement of \(\goal\) regions. This option is a good choice if \(k << 2^{\vert \goal \vert}\). The value of \(k\) is based on domain knowledge of the environment of interest. In our experiments, we use Option \ref{it:ps-opt1} or relax the assumption to show that it often works in practice without it.

\end{document}